%% file: main.tex
\documentclass{article}

\usepackage[preprint,nonatbib]{neurips_2023}

\usepackage[utf8]{inputenc} 
\usepackage[T1]{fontenc}    
\usepackage{hyperref}       
\usepackage{url}            
\usepackage{booktabs}       
\usepackage{amsfonts}       
\usepackage{nicefrac}       
\usepackage{microtype}      
\usepackage[dvipsnames]{xcolor}
\usepackage{titlesec}
\usepackage{tcolorbox}
\usepackage{mdframed}
\usepackage{makecell}
\usepackage{enumitem}
\usepackage{graphicx}
\usepackage{subcaption}
\usepackage{pifont}
\usepackage{url}
\usepackage{times}
\usepackage{latexsym}
\usepackage[T1]{fontenc}
\usepackage[utf8]{inputenc}
\usepackage{microtype}
\usepackage{inconsolata}
\usepackage{alltt}
\usepackage{multirow}
\usepackage{tcolorbox}
\usepackage{multicol,multirow}
\usepackage{color,colortbl}
\usepackage{stfloats}
\usepackage{booktabs}
\usepackage{xspace}
\usepackage{listings}
\usepackage{makecell}
\usepackage{colortbl}
\usepackage{mdframed}
\usepackage{amssymb}  
\usepackage{amsthm}
\usepackage{amsmath}
\usepackage{booktabs}
\usepackage{subcaption}
\usepackage{cleveref}
\usepackage{appendix}
\usepackage{algorithm2e}
\usepackage{hyperref}

\SetKwComment{Comment}{\color{green!50!black}// }{}

\SetKwProg{Function}{def}{}{}

\newcommand{\na}{{n/a}}

\definecolor{lightblue}{RGB}{173,216,230}
\definecolor{lightcoral}{RGB}{240,128,128}
\definecolor{lightgoldenrodyellow}{RGB}{250,250,210}
\definecolor{lightgreen}{RGB}{144,238,144}
\definecolor{lightpink}{RGB}{255,182,193}
\definecolor{lightsalmon}{RGB}{255,160,122}
\definecolor{lightseagreen}{RGB}{32,178,170}
\definecolor{lightskyblue}{RGB}{135,206,250}

\hypersetup{
    colorlinks=true,
    linkcolor=blue,
    filecolor=magenta,      
    urlcolor=lightgreen,
    pdftitle={Overleaf Example},
    pdfpagemode=FullScreen,
    }

\global
\tcbset{
  aibox/.style={
    width=\textwidth,
    top=10pt,
    colback=white,
    colframe=black,
    colbacktitle=black,
    center,
  }
}
\newtcolorbox{AIbox}[2][]{aibox,title=#2,#1}
\tcbset{
  aiboxsmall/.style={
    width=0.62\textwidth,
    top=10pt,
    colback=white,
    colframe=black,
    colbacktitle=black,
    enhanced,
    top,
    attach boxed title to top left={yshift=-0.1in,xshift=0.15in},
    boxed title style={boxrule=0pt,colframe=white,},
  }
}
\newtcolorbox{AIboxSmall}[2][]{aiboxsmall,title=#2,#1}
\definecolor{aigold}{RGB}{244,210, 1} 
\definecolor{aired}{RGB}{255,180,181}
\newlength\savewidth

\definecolor{defaultcolor}{gray}{0.9}

\newcommand{\ourmethod}{\textit{Set-of-Mark}}
\newcommand{\ourmodelfortitle}{Set-of-Mark Prompting}
\newcommand{\ourmodel}{\textit{Set-of-Mark Prompting}}

\title{\includegraphics[scale=0.035]{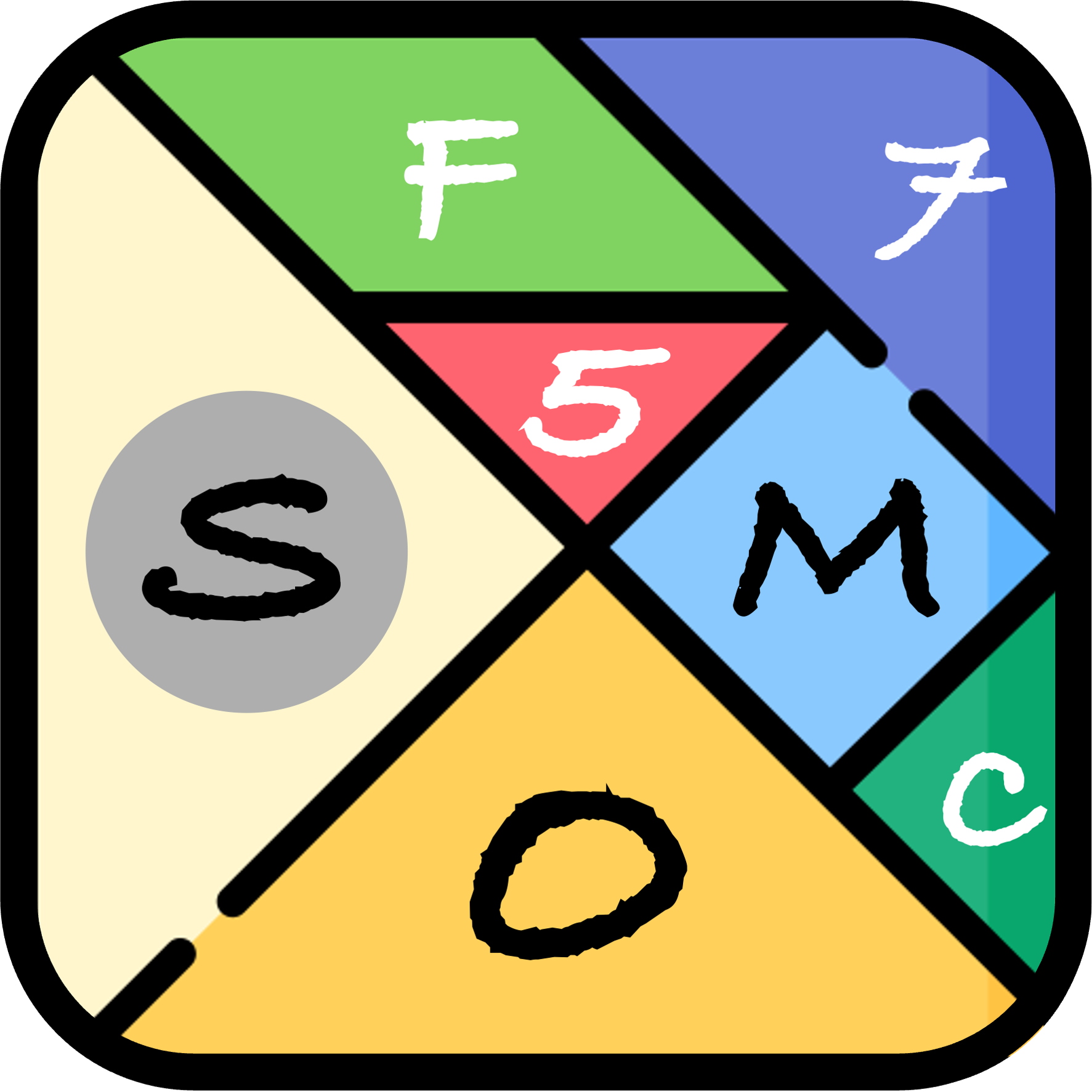} \ourmodelfortitle~Unleashes \\
Extraordinary Visual Grounding in GPT-4V}

\author{
\small{
 Jianwei Yang$^{1*\ddagger}$, ~Hao Zhang$^{2*}$, ~Feng Li$^{2*}$, ~Xueyan Zou$^{3*}$, ~Chunyuan Li$^{1}$, ~Jianfeng Gao{$^1$}
}
\and
{
\footnotesize
$^1$ Microsoft Research, Redmond \;  
$^2$ HKUST \;
$^3$ University of Wisconsin-Madison \;
}
\and
\footnotesize{
$^*$Core Contributor \;
$^{\ddagger}$~Project Lead \;
}
}
 
\begin{document}

\maketitle

\vspace{-30pt}
\begin{center}
\footnotesize{
\includegraphics[scale=0.02]{figures/som_logo.png} Playground: \url{https://som-gpt4v.github.io/}}
\end{center}
\vspace{-10pt}

\begin{figure}[h]
    \centering
    \includegraphics[width=0.96\textwidth]{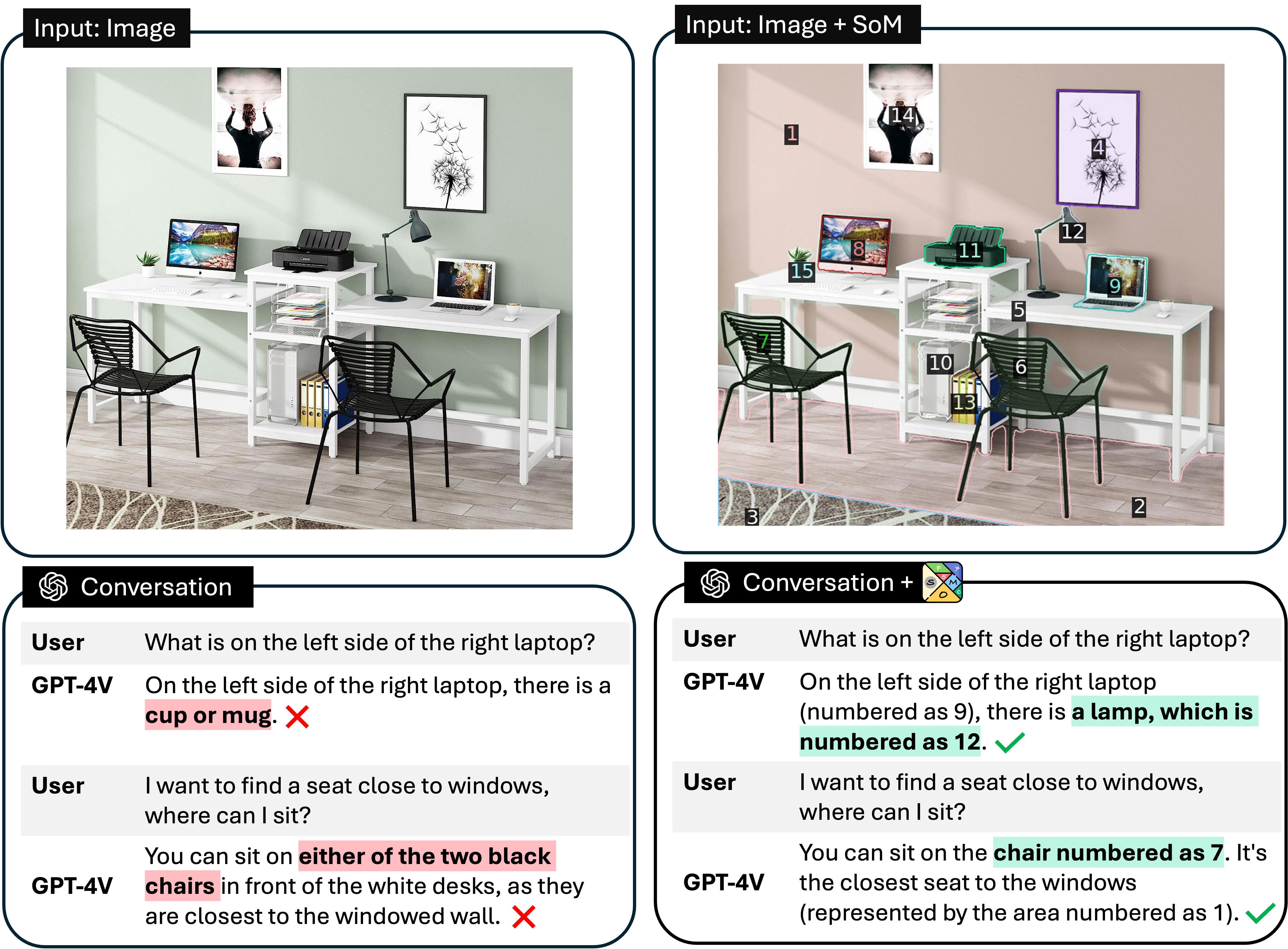}
    \vspace{-5pt}
    \caption{Comparisons of GPT-4V prompting techniques: (left) standard prompting and (right) the proposed \ourmodel. Simply overlaying ids on image regions unleashes visual grounding and corrects answers for GPT-4V. \emph{Note that no marks are leaked to user text inputs}..}
    \label{fig:teaser-comparison}
\end{figure}

\begin{abstract}

We present \emph{Set-of-Mark (SoM)}, a new visual prompting method, 
to unleash the visual grounding abilities of large multimodal models (LMMs), such as GPT-4V. 
As illustrated in Fig.~\ref{fig:teaser-comparison} (right), we employ off-the-shelf interactive segmentation models, such as SEEM/SAM, to partition an image into regions at different levels of granularity, and overlay these regions with a set of marks~\textit{e.g.,} alphanumerics, masks, boxes. Using the marked image as input, GPT-4V can answer the questions that require visual grounding.
We perform a comprehensive empirical study to validate the effectiveness of SoM on a wide range of fine-grained vision and multimodal tasks.
For example, our experiments show that GPT-4V with SoM outperforms the state-of-the-art fully-finetuned referring expression comprehension and segmentation model on RefCOCOg in a zero-shot setting. Code for \emph{SoM} prompting is made public here: \url{https://github.com/microsoft/SoM}.
\end{abstract}

\section{Introduction}

In the past few years, we have witnessed a significant advancement in large language models (LLMs)~\cite{brown2020language,bubeck2023sparks,chowdhery2022palm,touvron2023llama,zhang2022opt,openai2023gpt4}. In particular, Generative Pre-trained Transformers or GPTs~\cite{brown2020language,openai2023gpt4} have led a number of breakthroughs in the industry and academia. Since the release of GPT-4, large multimodal models (LMMs) have gained a growing interest in the research community. A number of works strive to build multimodal GPT-4 based on open-sourced models~\cite{liu2023visual,zhu2023minigpt,zhang2023llama,dai2023instructblip,wu2023visual,yang2023mm,li2023multimodal}.  
Recently, GPT-4V(ision) is released and attracts immediate attention from the community for its outstanding multimodal perception and reasoning capability. Its superiority and generality are showcased in~\cite{yang2023dawn}. 
Despite the unprecedented, strong, vision-language understanding capabilities, GPT-4V's fine-grained visual grounding ability is relatively weak, or yet to be \emph{unleashed}. For example, GPT-4V can hardly produce a sequence of accurate coordinates for a dog in the image\footnote{\url{https://blog.roboflow.com/gpt-4-vision/}}, or a few traffic lights~\cite{you2023ferret}. 

Our study is motivated by addressing the failed cases of GPT-4V on fine-grained vision tasks, such as object referring. In principle, these tasks require the model to have rich semantic understanding and accurate spatial understanding of visual contents. However, prompting GPT-4V to produce a sequence of tokens that contains textual description and numeric coordinates not only  hurts the natural fluency in LLMs but also dismisses the spatial understanding ability in vision models used by LMMs, as pointed out by ~\cite{chen2022pix2seq}. Therefore, in this study we focus on improving visual inputs by proposing a new visual prompting method to unleash the visual grounding capability of LMMs. 
Visual prompting has been explored for various vision and multimodal tasks~\cite{chen2022focalclick,kirillov2023segment,zou2023segment,wang2023seggpt,shtedritski2023does}. These works can be categorized into two categories. 
One is to encode visual prompts such as point, box, and stroke into latent features, which are then used to prompt the vision models~\cite{zou2023segment,kirillov2023segment}. 
The other is to overlay visual marks directly onto the input images. The overlaid marks could be a red circle~\cite{shtedritski2023does}, a highlighted region~\cite{yang2023finegrained}, or a few circles with arrows~\cite{yang2023dawn}. Although these studies demonstrate the promise of pixel-level visual prompting, they are limited to visually referring to one or a few objects. Moreover, all these marks are not easily ``speakable'' by LLMs, thus impeding the grounding capability for GPT-4V.

In this paper we present a new prompting mechanism called \emph{Set-of-Mark (SoM) prompting} , \textit{i.e.} simply adding a set of visual marks on top of image regions. We first partition an input image into a set of semantically meaningful regions. Then, we overlay each region with a mark of various formats such as numbers, alphabets, masks, or boxes. We perform an empirical study to validate whether GPT-4V could better ground the regions given these interpretable and ``spreadable'' visual marks. 
Our experiments show that SoM can drastically unleash the grounding capability of GPT-4V. 
As shown in Fig.~\ref{fig:teaser-comparison}, 
compared to the standard prompting, SoM helps GPT-4V accurately ground its answers in corresponding regions, \textit{e.g.}, the right laptop is ``9'' and the lamp is ``12''. Likewise, it can also associate the chair on the left with ``7''. Note that no information about the marks is leaked to the conversation with GPT-4V, which indicates that the model can not only understand the semantics of the visual marks but also know how and when to associate visual contents with marks. To the best of our knowledge, this is the first study to demonstrate that the emergent visual grounding ability of GPT-4V can be unleashed by visual prompting, SoM. To summarize, our main contributions are:
\begin{itemize}[leftmargin=*]
    \item We propose a simple yet effective visual prompting technique, called \emph{Set-of-Mark (SoM)} prompting for GPT-4V, and show empirically that SoM effectively unleashes the extraordinary visual grounding ability of GPT-4V.
    \item We have developed a new suite of evaluation benchmarks to examine the grounding ability of GPT-4V and other LMMs. For each image in the dataset, we employ off-the-shelf segmentation models to segment an image into regions, and overlay each region with visual marks, such as numeric numbers. 
    \item We have conducted quantitative and qualitative analysis to validate the effectiveness of SoM on a wide range of vision tasks. Our experiments show that SoM significantly improves GPT-4V's performance on complex visual tasks that require grounding. For example,  GPT-4V with SoM outperforms the state-of-the-art fully-finetuned referring segmentation model on RefCOCOg in a zero-shot setting.
\end{itemize}

\section{\ourmodelfortitle}

\begin{figure}
    \centering    
    \includegraphics[width=\textwidth]{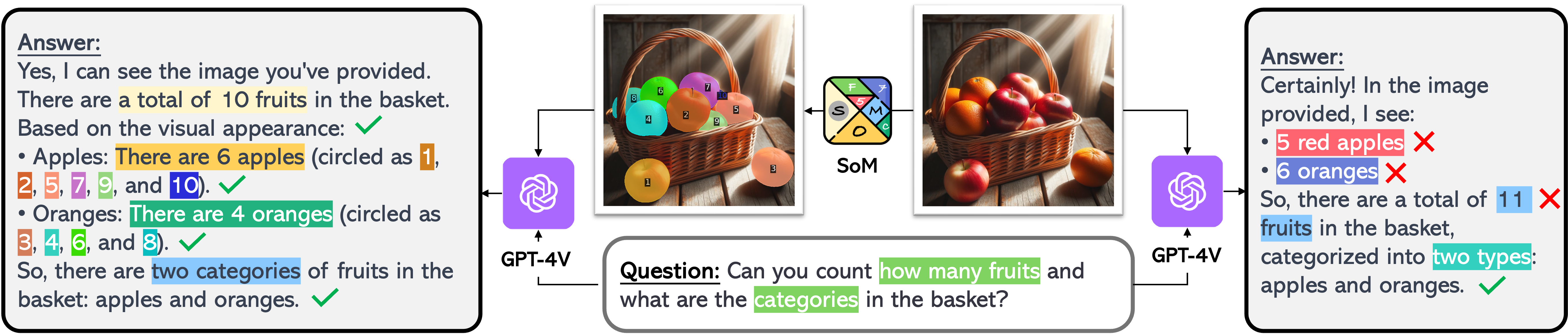}
    \caption{Comparing standard GPT-4V and its combination with \ourmethod~\textit{(SoM)} \textit{Prompting}. it clearly shows that our proposed prompting method helps GPT-4V to see more precisely and finally induce the correct answer. We highlight the differences between our method and the standard one. (The image is generated by Dalle-3 and is better viewed in color.)}
    \label{fig:GPT-4V_plus_SoM}
\end{figure}

This section introduces \ourmodel~and explains how it can be applied to prompt the LMMs, GPT-4V in particular. 

\subsection{Problem Definition}

Typically, LMMs $\mathcal{F}$ take an image $I \in \mathcal{R}^{H \times W \times 3}$ and a text query of length of $l_i$, $T^i = [t^i_1,...,t^i_{l_i}]$ as input, and generate a sequence of textual output with length $l_o$, $T^o = [t^o_1,...,t^o_{l_o}]$, formulated as: 
\begin{equation}
T^o = \mathcal{F}(I, T^i)    
\label{Eq:GPT-4V}
\end{equation}
Given the versatility of current LLMs used in LMMs, the input and output texts can be comprised of different types of characters (\textit{e.g.}, both alphabetic and numeric), and also multilingual. A large number of works have attempted to improve prompt engineering on the input text $T^i$ to endow more reasoning capability in LLMs~\cite{wei2022chain,yao2023tree}.

In contrast to prompt engineering for LLMs, the goal of this study is to develop a new prompting method for input images to unleash visual grounding ability of LMMs. In other words, we strive to equip a LMM with the ability to \textit{see location-by-location}. This necessitates two essential properties for the prompting strategy:
\begin{itemize}[leftmargin=*]
    \item {The method should be able to partition an image into a set of semantically meaningful regions to align with the textual outputs, an ability known as \emph{grounding}.}
    \item {The auxiliary information cast to the input image should be both \emph{interpretable} and \emph{speakable} by the LMM, so that it can be described in its textual outputs.}
\end{itemize}

With this in mind, we develop \ourmodel, a simple prompting method of overlaying a number of marks to the meaningful regions in an image. This operation augments the input image $I$ to a marked image $I^m$ while keeping the other inputs to LMMs unchanged, as shown in Fig.~\ref{fig:GPT-4V_plus_SoM}. Mathematically, Eq.~\eqref{Eq:GPT-4V} becomes 
\begin{equation}
    T^o = \mathcal{F}(\underbrace{\textit{SoM}(I)}_{I^m}, T^i).
\end{equation}

Although it is straightforward to apply SoM to all LMMs, we find that not all LMMs have the ability to ``speak out'' about the marks. Actually, we find that only GPT-4V, when equiped with SoM, shows the emergent grounding ability and significantly outperforms the other LMMs. 
In what follows, we will explain how we partition an image into regions and mark the image regions in SoM.

\subsection{Image Partition}

Given an input image, we need to extract meaningful and semantically aligned regions. Ideally, the extraction should be automatic or semi-automatic to avoid extra burden on users. To achieve this, we employ a suite of image segmentation tools. To support different use cases, the segmentation tools need to possess the following properties:
\begin{itemize}[leftmargin=*]
    \item \textbf{Strong performance}: the region partition should be precise enough to convey fine-grained spatial layout information to GPT-4V. In this sense, we choose one of the state-of-the-art image segmentation models MaskDINO~\cite{li2023mask}.
    \item \textbf{Open vocabulary}: the segmentation models should be open-vocabulary so that they can recognize objects out of predefined vocabulary. As such, we opt for the advanced models like SEEM~\cite{zou2023segment}. 
    \item \textbf{Rich granularity}: Finally, the region of interest might be not only a full object but also a part of it. Therefore, we also employ SAM~\cite{kirillov2023segment} and Semantic-SAM~\cite{li2023semanticsam}.
\end{itemize}

We have collected a suite of image segmentation models that offers a comprehensive toolbox for a user to partition arbitrary images. Most of models are interactive and promotable such that users can customize the SoM prompts interactively. 

\begin{figure}
    \centering
    \includegraphics[width=\textwidth]{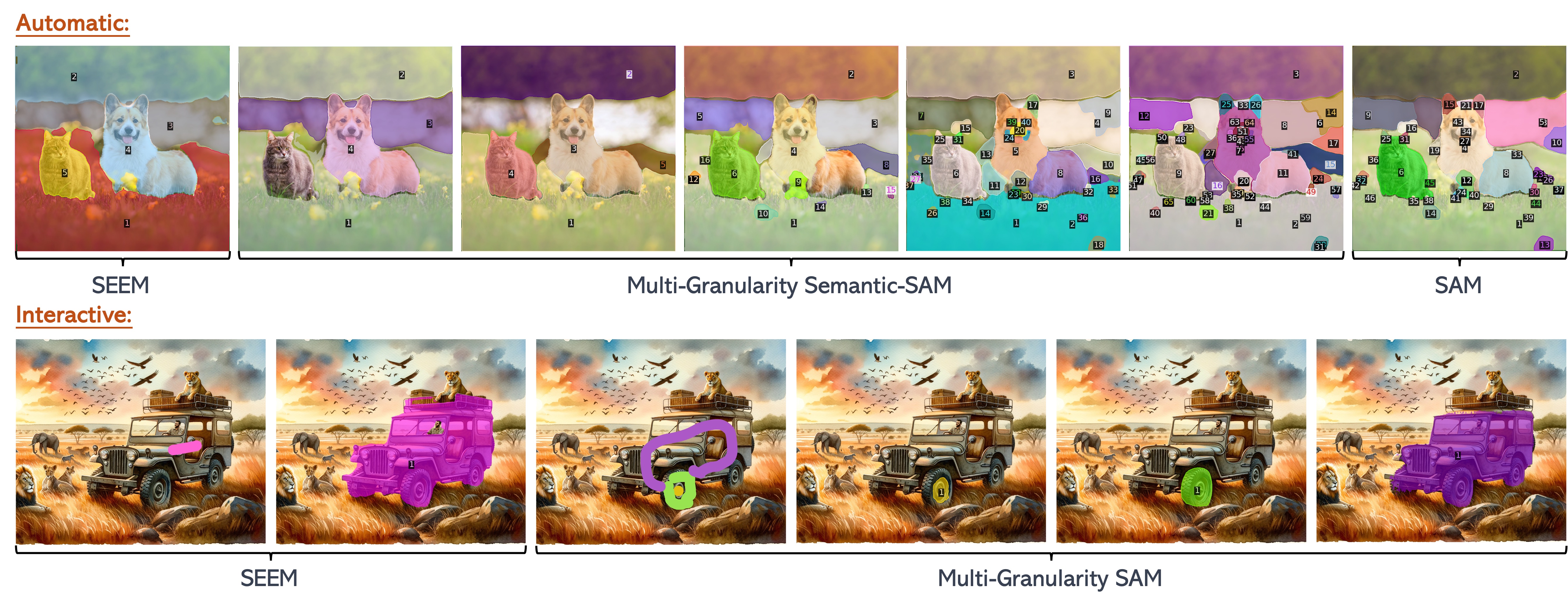}
    \caption{We compile different segmentation models including SEEM~\cite{zou2023segment}, Semantic-SAM~\cite{li2023semanticsam} and SAM~\cite{kirillov2023segment} as the image partition tools used in our work. Users can select which granularity of masks to generate, and which mode to use between automatic (top) and interactive (bottom). A higher alpha blending value (0.4) is used for better visualization.}
    \label{fig:image-partition}
\end{figure}

Based on our image partition toolkit, we divide an input image $I$ of size $H \times W$ into $K$ regions $R = [r_1, ..., r_K] \in \{0,1\}^{K \times H \times W}$, which are represented by $K$ binary masks. We show some SoM examples in Fig.~\ref{fig:image-partition} generated using our toolbox.

\subsection{Set-of-Mark Generation}

Once we obtain image partition $M$, we need to generate for each region a mark that is useful for grounding by GPT-4V. We consider two factors, the types and locations of marks.

\begin{figure}[t]
    \centering
    \includegraphics[width=\textwidth]{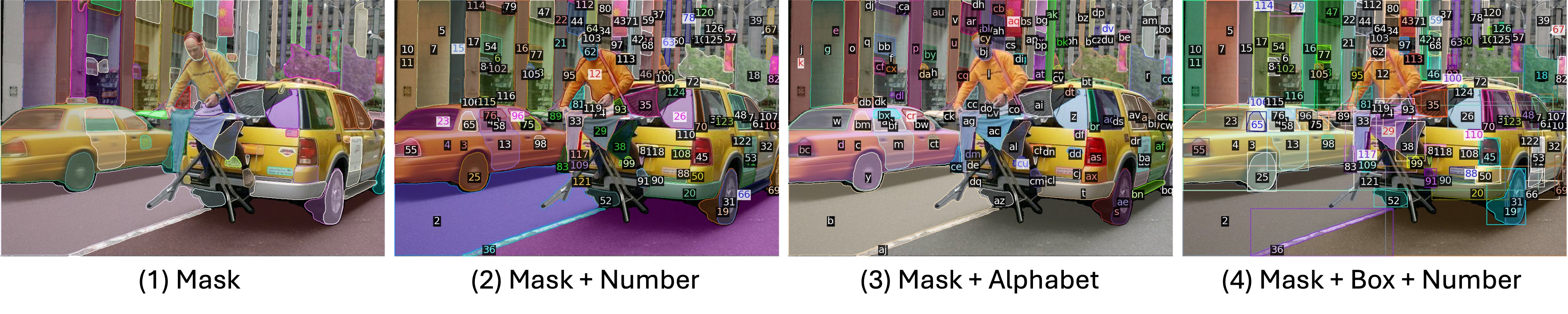}
    \caption{Different types of marks can be used in our \ourmethod.}
    \label{fig:mark-types}
\end{figure}

\paragraph{Mark Type.} As we discussed earlier, the mark type depends on whether they can be interpreted by GPT-4V. In this work, we consider alphanumeric in that it is not only compact enough to not occupy much image space but recognizable by GPT-4V (using its OCR capability). Besides, we also consider boxes and mask boundaries as auxiliary marks. In addition, we note that the mark types should be image-dependent to avoid any conflicts with the original image contents. For example, given an arithmetic image full of numbers, the numeric marks should be avoided, while the alphabetic characters should be not used for a screenshot of a document. We leave the automatically determining which mark types to use and how to combine them to future work. Fig.~\ref{fig:mark-types} shows different mark types and the interpreted outputs from GPT-4V.

\begin{figure}[t]
\begin{minipage}{.45\textwidth}
  \includegraphics[width=1.0\linewidth]{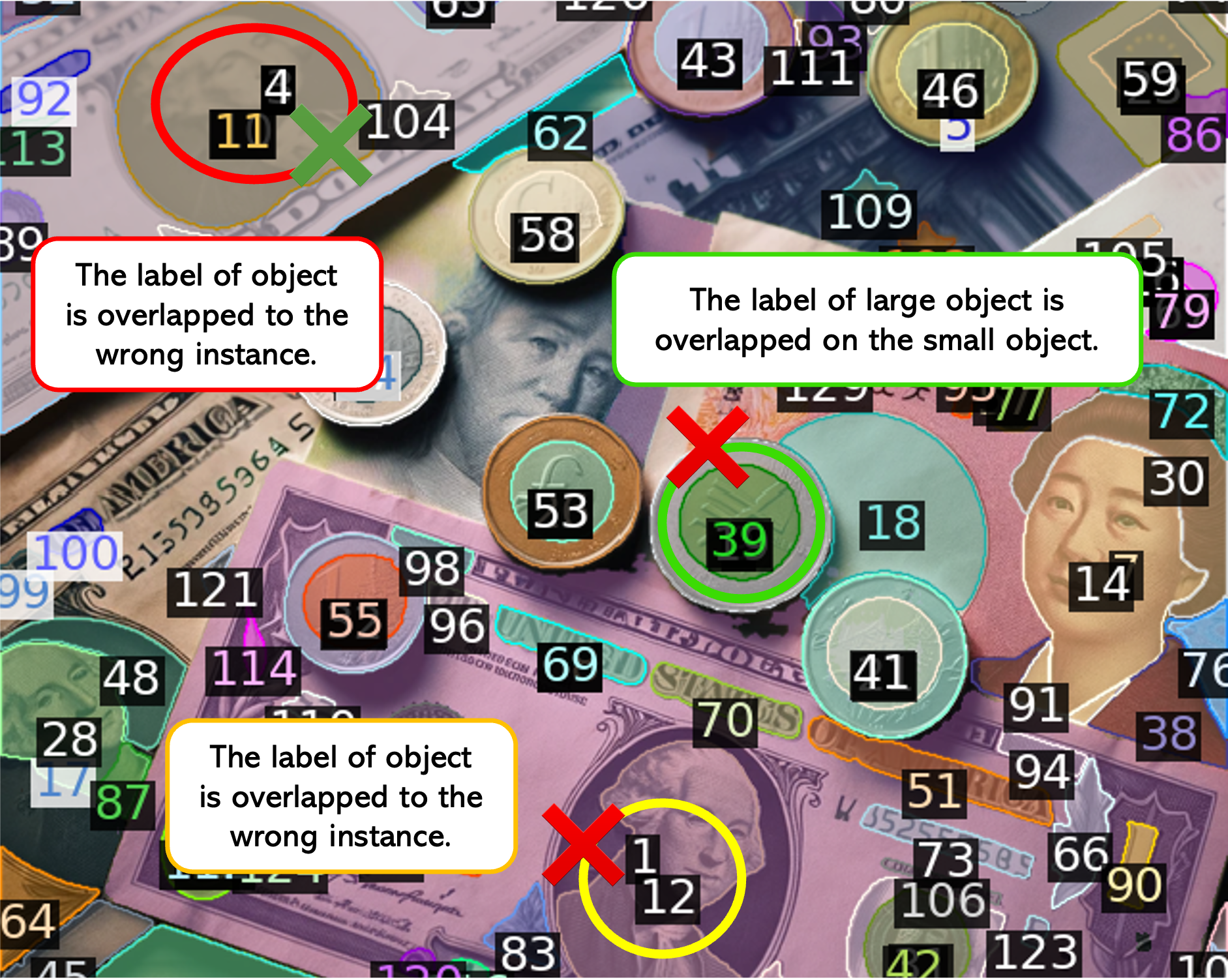}
\end{minipage}
\hfill
\begin{minipage}{.52\textwidth}
  \includegraphics[width=1.0\linewidth]{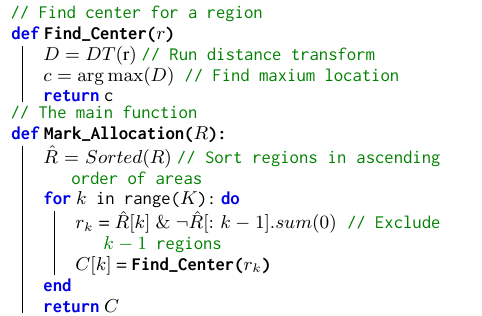}
\end{minipage}%
\vspace{-5pt}
\caption{Left: some conflicts caused by putting all marks in the center. Right: our proposed mark allocation algorithm to address the conflicts.}   
\label{fig:mark_failure} 
\end{figure}

\paragraph{Mark Location.}
Given an image of size $H \times W$, we need to find good mark locations for all $K$ regions. A straightforward way is to derive the center coordinate $c_k=(x_k^c, y_k^c)$ of $k$-th mask, and overlay the marks there. However, this inevitably introduces some overlaps or conflicts that may confuse GPT-4V, especially for images with dense objects, e.g., two objects centered around similar locations or concave regions, as shown in Fig.~\ref{fig:mark_failure}. 

\begin{figure}[!ht]
\centering
    \begin{AIbox}{SoM-Enabled Plain and Interleaved Prompts}
\begin{minipage}{0.48\textwidth}
    \includegraphics[width=1.0\textwidth]{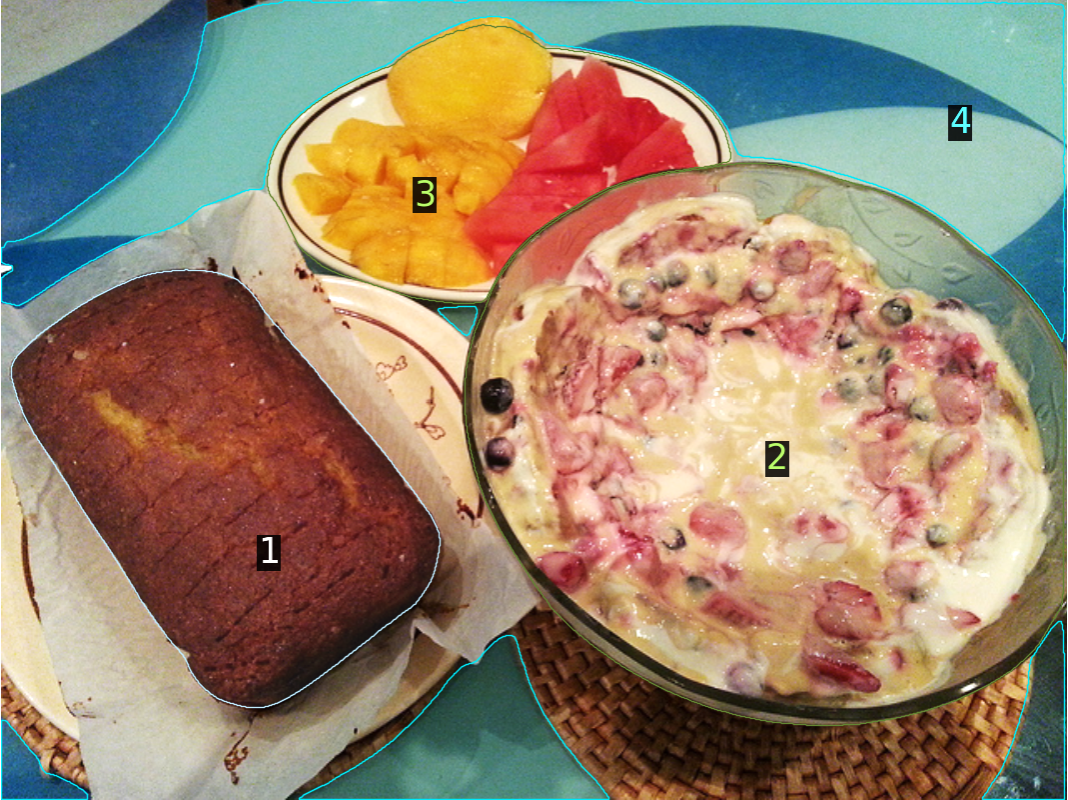}
    \\
    \\
    \footnotesize
    \textbf{\textcolor{Orange}{User}}: What is in the image?
    \\
    \textbf{\textcolor{Purple}{GPT-4V}}: In the image, I observe the following items:

    \colorbox{yellow}{\textbf{Loaf or Cake ({1})}}: A rectangular-shaped baked item, possibly a loaf of bread or a cake.

    \colorbox{lightblue}{\textbf{Creamy Fruit Mixture ({2})}}: A bowl containing a creamy mixture with various berries and possibly other fruits. It might be a fruit salad with a creamy dressing, yogurt with fruits, or some kind of dessert.
    \\
    \colorbox{lightpink}{\textbf{Sliced Fruits ({3})}}: A plate with slices of various fruits, including what looks like pineapples and possibly slices of watermelon or another type.
    \\
    \colorbox{lightgreen}{Table Surface (4)}: A portion of the table or countertop with a bluish hue.

\end{minipage}    
\hfill
\begin{minipage}{0.48\textwidth}
    \footnotesize
    (con't) Additionally, there are various decorative elements, plates, and bowls.
    \\
    \\
    \textbf{\textcolor{Orange}{User}}: What is in \colorbox{lightpink}{3}?
    \\
    \textbf{\textcolor{Purple}{GPT-4V}}: In the image, the item labeled "\colorbox{lightpink}{3}" appears to be a plate of sliced fruits. It looks like there's pineapple and possibly another type of melon or fruit, such as watermelon or pink grapefruit.   
    \\
    \\
    \textbf{\textcolor{Orange}{User}}: Can I put the fruits in \colorbox{lightpink}{3} into \colorbox{lightblue}{2}?
    \\
    \textbf{\textcolor{Purple}{GPT-4V}}: Yes, based on the image, the fruits in \colorbox{lightpink}{3} appear to be sliced fresh fruits, and they can be mixed or poured into the creamy fruit mixture in \colorbox{lightblue}{2} to enhance flavor and texture. It's a common practice in desserts to combine fresh fruits with creamy mixtures.    
    \\
    \\
    \textbf{\textcolor{Orange}{User}}: Which can be toasted, \colorbox{yellow}{1} or \colorbox{lightblue}{2}?
    \\
    \textbf{\textcolor{Purple}{GPT-4V}}: Based on the image, item \colorbox{yellow}{1} (which appears to be a loaf of bread or cake) is the one that can be toasted. Item \colorbox{lightblue}{2} is a creamy fruit mixture and is not suitable for toasting.
    \\
    \\
    \textbf{\textcolor{Orange}{User}}: How long it usually takes to toast a slice of loaf in \colorbox{yellow}{1}?
    \\
    \textbf{\textcolor{Purple}{GPT-4V}}: Typically, to toast a slice of a loaf like the one labeled as "\colorbox{yellow}{1}" in the image, it would take between 2 to 5 minutes in a toaster on a medium setting. 
    \\
    \\
    \end{minipage}
    \end{AIbox}   
\caption{We can use two types of text prompts to LLMs: Left shows a plain text prompt, which does not include any clues about the marks; Right is an interleaved text prompt, which uses the marks to ground image regions.}
\label{fig:LLMs_prompt}
\end{figure}

To mitigate the problem, we propose a mark allocation algorithm as illustrated in the algorithm on the right side of Fig.~\ref{fig:mark_failure}. Given the set of masks $M$, we first calculate the areas for all and sort them in an ascending order (line 6). This strategy ensures that smaller regions are considered before  large regions. To further avoid the potential overlaps, for $k$-th mask, we exclude the region that is covered by any $k-1$ masks (line 8). The resulting mask is then fed to a distance transform algorithm, which helps to find the location inside the mask where the minimal distance to all boundary points is maximal. In practice, however, a region may be so small that the mark could cover the (almost) whole region. In this case, we move the marks off the region slightly. We find that GPT-4V can still build a decent association between the marks and regions.

Once we determine the mark type and locations for all regions, we overlay mark $m_k$ to region $r_k$ at location $c_k$. We make sure that each mark is unique so that the full set $M = \{m_1,...m_k\}$ are distinguishable and speakable by LLMs. 

\subsection{Interleaved Prompt}

Thus far we have obtained a new image $I^m$ with overlaid marks. Additionally, we have a set of $K$ region-mark pairs $\{\langle r_1,m_1 \rangle, ..., \langle r_K, m_K \rangle \}$. Given the additional cues in $I^m$, we can use either a plain text prompt or interleaved ones to prompt LMMs:

\begin{itemize}[leftmargin=*]
    \item \textbf{Plain Text Prompt}. We can use plain text prompts as usual without any explicit reference to the marks/regions in the image. As shown in Fig.~\ref{fig:teaser-comparison}, even without any special textual prompt, GPT-4V can automatically ground itself in the regions and corresponding marks. It can be used in a wide range of scenarios where users do not have specific regions of interest. 
    \item \textbf{Interleaved Text Prompt}. We can use interleaved ones by injecting the marks into the text directly. Since the marks are interpretable to LMMs, we can seamlessly blend them into the original text prompt to make a symbolic reference. 
\end{itemize}

Examples of applying these two types of text prompts based on SoM are demonstrated in Fig.~\ref{fig:LLMs_prompt}. Note that for each question, we use a new chat window to avoid context leakage during the conversation. In real-world scenarios, the above two text prompting strategies can be combined and used in multi-round conversations with GPT-4V to significantly enrich human-AI interaction. Moreover, users can also opt to draw the marks by themselves or revise the marks generated using the toolbox. 

\section{SoM Prompting for Vision}

\begin{figure}[t]
\centering


\begin{minipage}{0.48\textwidth}
    \begin{AIbox}{Open-Vocabulary Segmentation}
    \includegraphics[width=1.0\textwidth]{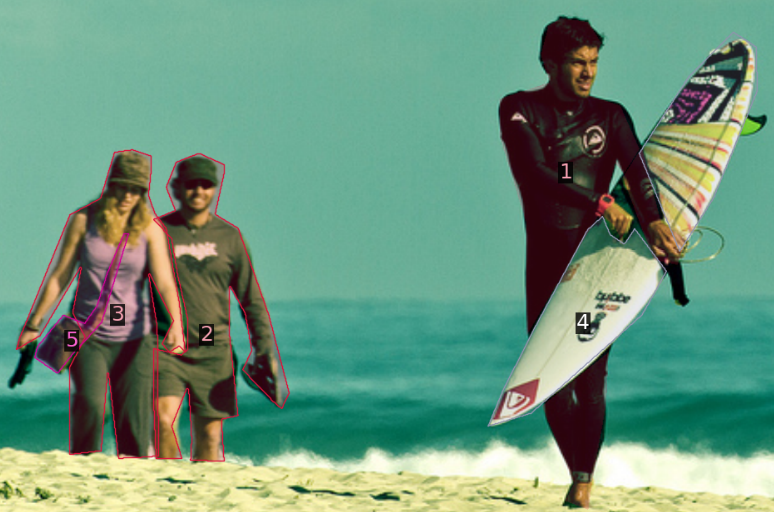}
    \\
    \footnotesize
    \textbf{\textcolor{Orange}{User}}: I have labeled a bright numeric ID at the center for each visual object in the image. Please enumerate their names. You must answer by selecting from the following names: [Person, ..., curtain] (ignore the 80 categories in COCO)

    \textbf{\textcolor{Purple}{GPT-4V}} 1. Person 2. Person 3. Person 4. Surfboard 5. Handbag

    \end{AIbox}   
\end{minipage}
\begin{minipage}{0.48\textwidth}
    \begin{AIbox}{Referring Segmentation}
    \includegraphics[width=1.0\textwidth]{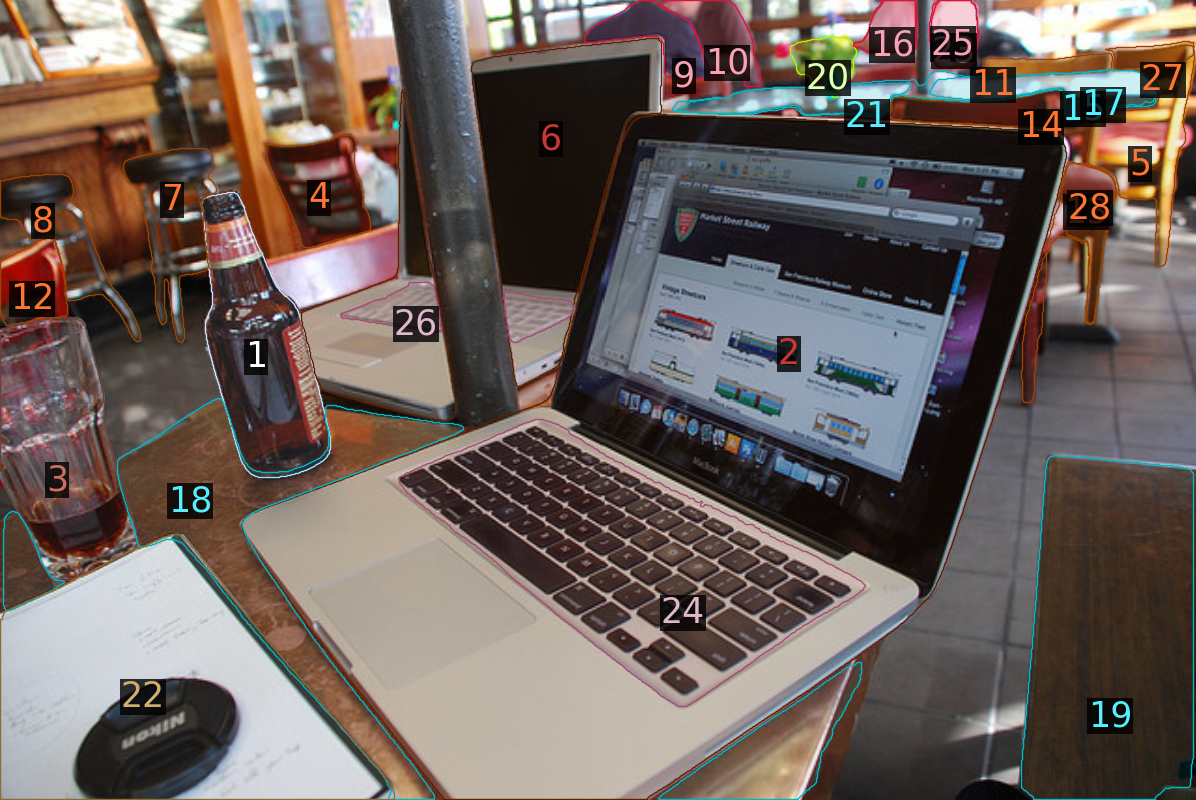}
    \\
    \footnotesize
    \textbf{\textcolor{Orange}{User}}: I have labeled a bright numeric ID at the center for each visual object in the image. Please tell me the IDs for: The laptop behind the beer bottle; Laptop turned on.

    \textbf{\textcolor{Purple}{GPT-4V}} The IDs for the items you've mentioned: The laptop behind the beer bottle: 6
Laptop turned on: 2
\\
    \end{AIbox}   
\end{minipage}
\begin{minipage}{0.48\textwidth}
    \begin{AIbox}{Video Object Segmentation}
    \includegraphics[width=1.0\textwidth]{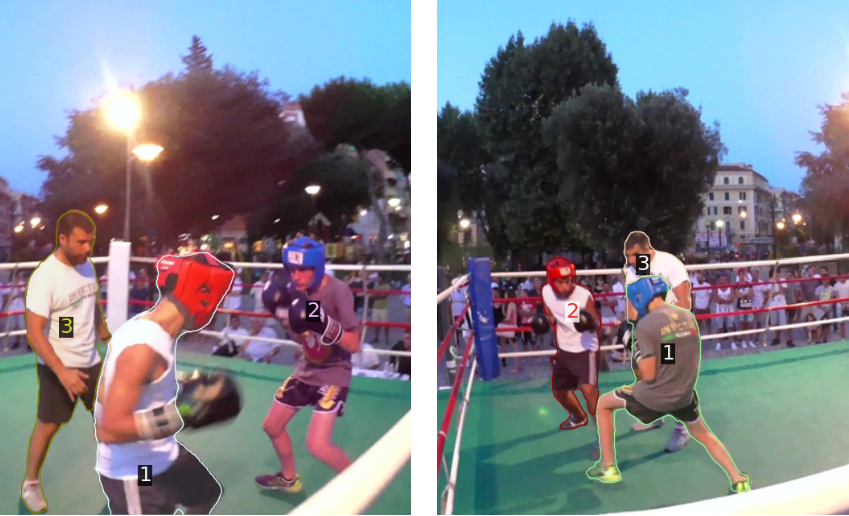}
    \\
    \footnotesize
    \textbf{\textcolor{Orange}{User}}: The 2 images are from the same video, where the first image is the first frame and the second image is a later frame. In the first image, there are 2 objects labeled with 1,2. Can you track these 2 objects in the second image? 

    \textbf{\textcolor{Purple}{GPT-4V}} 1. The object labeled with 1 (a boxer in red headgear) is most similar to the object labeled with 2 (another boxer in red headgear).
2. The object labeled with 2 (a boxer in blue) is most similar to the object labeled with 1 (another boxer in blue).
    \end{AIbox}   
\end{minipage}
\begin{minipage}{0.48\textwidth}
    \begin{AIbox}{Grounded Segmentation}
    \includegraphics[width=0.95\textwidth]{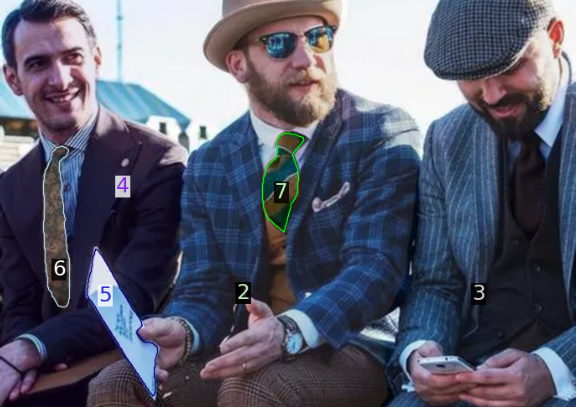}
    \\
    \footnotesize
    \textbf{\textcolor{Orange}{User}}: I have labeled a bright numeric ID at the center for each visual object in the image. Given the image showing a man in glasses holding a piece of paper, find the corresponding regions for a man in glasses, a piece of paper. 

    \textbf{\textcolor{Purple}{GPT-4V}} The "man in glasses" corresponds to the region labeled with "2". The "piece of paper" corresponds to the region labeled "5".
    \\
    \\

    \end{AIbox}   
\end{minipage}
\caption{Examples of some vision tasks we benchmark on GPT-4V.}
\vspace{-14pt}
\label{fig:vision_tasks}
\end{figure}

We highlight that the unique merit of using SoM to prompt GPT-4V is that it can produce outputs beyond texts. Since each mark is exclusively associated with an image region represented by a mask, we can trace back the masks for any mentioned marks in the text outputs. Consider the example in Fig.~\ref{fig:vision_tasks}, the GPT-4V response on the left side contains the names and details of four regions. It can induce the one-to-one mappings between marks and text description, \textit{i.e.}, $m_k \leftrightarrow text_k$. Given that $r_k \leftrightarrow m_k$, we can further associate textual description with the masks for all regions. Finally, we can bridge the triplets $\langle r_k, m_k, text_k\rangle$ for all regions, \textit{i.e.}, $r_k \leftrightarrow m_k \leftrightarrow text_k$. The ability to produce paired texts and masks allows the \textit{SoM} prompted GPT-4V to produce plausible visually grounded texts and more importantly support a variety of fine-grained vision tasks, which are challenging for the vanilla GPT-4V model.

\paragraph{Vision Tasks.}  We quantitatively examine the performance of \textit{SoM} prompted GPT-4V. With simple prompt engineering, it can be readily used in, but not limited to, a wide range of vision tasks such as
\begin{itemize}[leftmargin=*]
    \item \textbf{Open-vocabulary Image Segmentation}: We ask GPT-4V to exhaustively tell the categories for all marked regions and the categories that are selected from a predetermined pool.
    \item \textbf{Referring Segmentation}: Given a referring expression, the task for GPT-4V is selecting the top-matched region from the candidates produced by our image partition toolbox. 
    \item \textbf{Phrase Grounding}: Slightly different from referring segmentation, phrase grounding uses a complete sentence consisting of multiple noun phrases. We ask GPT-4V to allocate the corresponding regions for all labeled phrases.
    \item \textbf{Video Object Segmentation}: It takes two images as input. The first image is the query image which contains a few objects of interest to identify in the second image. Given that GPT-4V supports multiple images as input, our prompting method can also be applied to ground visual objects across frames in a video.
\end{itemize}

We show how we prompt GPT-4V using SoM for the above vision tasks in Fig.~\ref{fig:vision_tasks}. It is worth noting that SoM can be applied to broader tasks, such as region captioning, keypoint localization, and part segmentation, \textit{etc}. Due to the limited access (quota) to GPT-4V, we focus on the aforementioned vision tasks and leave the explorations to other tasks for future work. We present the detailed empirical study in the next section.

\section{Experiments}

\subsection{Experimental Setup}

\paragraph{Implementation.} We do not need to train any models for our method. However, due to the limited quota and absence of GPT-4V API, we have to exhaustively send the \textit{SoM} augmented images to the ChatGPT interface. The authors in this work used a divide-and-conquer strategy to run the experiments and evaluations. For each instance, we use a new chat window so that there is no context leakage during the evaluation. In this sense, all the reported results for our method are zero-shot. Unless otherwise mentioned, we always use the numeric numbers as the marks for reporting the main results. The usage of other mark types is shown in our ablation and qualitative studies.

\paragraph{Benchmarks.} Given the limited quota of GPT-4V, it is not possible for now to fully evaluate the validation set in each individual vision task as we listed above. Consequently, we select a small subset of validation data from each dataset for our study. For each image in the datasets, we overlay a set of marks on the regions extracted using our image partition toolbox. Depending on the specific task, we exploit different partition tools to propose regions. In Table~\ref{tab:visual_grounding_benchmark}, we list the setup for each task.

\paragraph{Comparisons.} We compare our method with:

\begin{itemize}[leftmargin=*]
    \item \textbf{GPT-4V baseline predicting coordinates}. We use this as our baseline model. By default, GPT-4V can neither predict boxes nor generate masks. Following prior practice, we prompt the model to predict box coordinates. Comparing with the default GPT-4V baseline helps to inspect the benefit of our proposed \ourmodel.
    \item \textbf{State-of-the-art specialist  models}. For each of the vision tasks, a number of methods have been proposed. We choose state-of-the-art and representative methods for comparison. More specifically, MaskDINO~\cite{li2023mask} for generic segmentation on COCO, OpenSeeD~\cite{zhang2023simple} for zeroshot segmentation on ADE20K, Grounding DINO~\cite{liu2023grounding} and GLIPv2~\cite{zhang2022glipv2} for phrase grounding on Flick30K, Grounding DINO and PolyFormer~\cite{liu2023polyformer} for referring expression comprehension, PolyFormer and SEEM~\cite{zou2023segment} for referring expression segmentation and SegGPT~\cite{wang2023seggpt} for video object segmentation. We attempt to shed light on the gap between the strongest generalist vision model GPT-4V and specialist models that are sophisticatedly designed and trained with the take-specific data.
    \item \textbf{Open-sourced LMMs}. We quantitatively evaluate one of the state-of-the-art open-sourced LMMs, LLaVa-1.5~\cite{liu2023improved}, and qualitatively compare with MiniGPT-v2~\cite{Chen2023minigpt} in our study. Note that both models are trained with a good amount of data from the targeted vision tasks. We are the first to compare closed- and open-sourced LMMs on vision benchmarks.
\end{itemize}

\begin{table}[]
    \centering
    \footnotesize
\resizebox{0.98\linewidth}{!}{        
    \begin{tabular}{lccccc}
    \toprule
    Vision Task & Dataset &  \#Images & \#Instances & Marks  & Metric \\
    \midrule
      Generic Segmentation   &  COCO~\cite{lin2015microsoft} & 100 & 567 & Number \& Mask & Precision \\
      OV Segmentation   &  ADE20K~\cite{zhou2017scene} & 100 & 488 & Number \& Mask & Precision \\
      Phrase Grounding  & Flickr30K~\cite{plummer2016flickr30k} & 100 & 274 & Number \& Box & Recall@1 \\
      Referring Expression Segmentation & RefCOCO~\cite{kazemzadeh2014referitgame} & 100 & 177 &  Number \& Mask & mIoU \\
      Referring Expression Comprehension & RefCOCO~\cite{kazemzadeh2014referitgame} & 100 & 177 &  Number \& Mask & ACC@0.5 \\
      Video Object Segmentation & DAVIS~\cite{perazzi2016benchmark} &  71 & 157 &Number \& Mask & J\&F \\
    \bottomrule
    \end{tabular}
    }
    \vspace{5pt}
    \caption{Benchmarks used to evaluate the visual grounding capability of GPT-4V. Note that we select a small subset of images or videos from the corresponding datasets.}
    \label{tab:visual_grounding_benchmark}
    \vspace{-20pt}
\end{table}

\subsection{Quantitative Results}

We present the experimental results in Table~\ref{tab:main_results}.

\input{tables/main_results}
\begin{table}[]
    \centering
    \footnotesize
    
    \begin{minipage}{0.48\textwidth}
        \centering
        \begin{tabular}{lc}
            \toprule
            Mark Type & Flickr30K (R@1) \\
            \midrule
            Number \& Mask & 84.4 \\
            Number \& Mask \& Box & 89.2 \\
            \bottomrule
        \end{tabular}
        \vspace{1ex} 
        \caption{The performance on Flickr30K with different mark types on a subset of our dataset.}
        \label{tab:ablation mark type1}
    \end{minipage}%
    \hfill
    \begin{minipage}{0.48\textwidth}
        \centering
        \begin{tabular}{lc}
            \toprule
            Mask type & Refcocog (mIoU) \\
            \midrule
            MaskDINO & 75.6 \\
            GT mask & 90.1 \\
            \bottomrule
        \end{tabular}
        \vspace{1ex} 
        \caption{The performance on Flickr30K with different mask types.}
        \label{tab:ablation mark type2}
    \end{minipage}
    \vspace{-20pt}
\end{table}

\paragraph{Image Segmentation}
On image segmentation tasks, we evaluate the classification precision of GPT-4V + SoM and compare it with the strong segmentation model MaskDINO on COCO Panoptic segmentation dataset and OpenSeeD on ADE20K Panoptic segmentation dataset. For MaskDINO and OpenSeeD, we give them GT boxes and evaluate the output class. For GPT-4V + SoM, we overlay GT masks with alpha=0.4 and 0.2 for COCO and OpenSeeD, respectively, and add the ID number on each mask. We provide GPT-4V with the vocabulary of the datasets and ask it select a class label for each region. The results show that zero-shot performance of GPT-4V + SoM is close to the performance of fine-tuned MaskDINO and is much higher than the zero-shot performance of OpenSeeD on ADE20K. The similar performance on COCO and ADE20K for GPT-4V indicates its strong generalization ability to a wide of visual and semantic domains.

\paragraph{Referring}
For referring tasks, we evaluate RES and REC on RefCOCOg. We use MaskDINO to propose masks and overlay the masks and numbers on the images. 
We use mIoU as the evaluation metric and compare it with state-of-the-art specialist PolyFormer~\cite{liu2023polyformer} and SEEM~\cite{zou2023segment}. Accordingly, GPT-4V + SoM outperforms PolyFormer by a large margin. Note that the performance of PolyFormer is evaluated on our dataset for apple-to-apple comparison. For REC, we convert the masks into boxes for our method. When prompting GPT-4V to directly output the coordinates, we attain significantly poor performance (25.7), which verifies our earlier hypothesis. Once augmented with SoM, GPT-4V beats both specialists such as Grounding DINO and Polyformer and recent open-source LMM including Shikra, LLaVA-1.5, MiniGPT-v2, and Ferret. 

\paragraph{Phrase Grounding}
For phrase grounding on Flickr30K, we use Grounding DINO to generate box proposals for each image. We use a threshold of 0.27 to filter redundant boxes, then we use SAM to predict a mask for each box. We draw boxes, masks, and numbers on the images. We give GPT-4V a caption and the noun phrases of the caption for each image and let GPT-4V to ground each noun phrase to a region. Our zero-shot performance is comparable with SOTA models GLIPv2 and Grounding DINO.

\paragraph{Video Object Segmentation} 
We evaluate DAVIS2017~\cite{pont20172017} for the video segmentation task. To get the mask proposals of each video frame, we use MaskDINO~\cite{li2023mask}. The predicted masks are then overlayed on the corresponding frames with numeric labels following our SoM. We follow the semi-supervised setting to use the first frame masks as the reference and segment all the other frames of the same video. As GPT-4V can take in multiple images, we prompt GPT-4V with the first frame and the current frame to do segmentation by comparing similar objects across images. As shown in the last column of Table~\ref{tab:main_results}, GPT-4V achieves the best tracking performance (78.8 J\&F) compared with other generalist vision models.
\subsection{Ablation Study}

\paragraph{Mark type.} We study how the choice of mark types affects the final performance for the phrase grounding task on Flickr30k. We compare two types of marks. The first is numbers and masks and the second is numbers, masks, and boxes. As shown in Table~\ref{tab:ablation mark type1}, the scheme of adding extra boxes can improve the performance significantly.



\paragraph{Golden mark location.}  The extracted regions from our toolbox may often come with some errors. In this study, we investigate how GPT-4V behaves when we generate the marks with ground-truth annotations. Specifically, we replace the predicted segmentation masks with ground-truth ones in our RefCOCOg validation set. This means GPT-4V only needs to select one from regions for annotated phrases. As expected, the performance for referring segmentation can be further improved, especially when the segmentation model has some missed regions. The results are shown in Table~\ref{tab:ablation mark type2}, which shows using ground-truth masks in our SoM improves the RefCOCOg performance by $14.5$ mIoU. We also observe that most failure cases when using GT masks are not the GPT-4V's problem, but the annotation itself is ambiguous or incorrect.



\subsection{Qualitative Observations}

When running through this work, we observed several intriguing qualitative findings.

\begin{figure}[t]
    \includegraphics[width=0.99\linewidth]{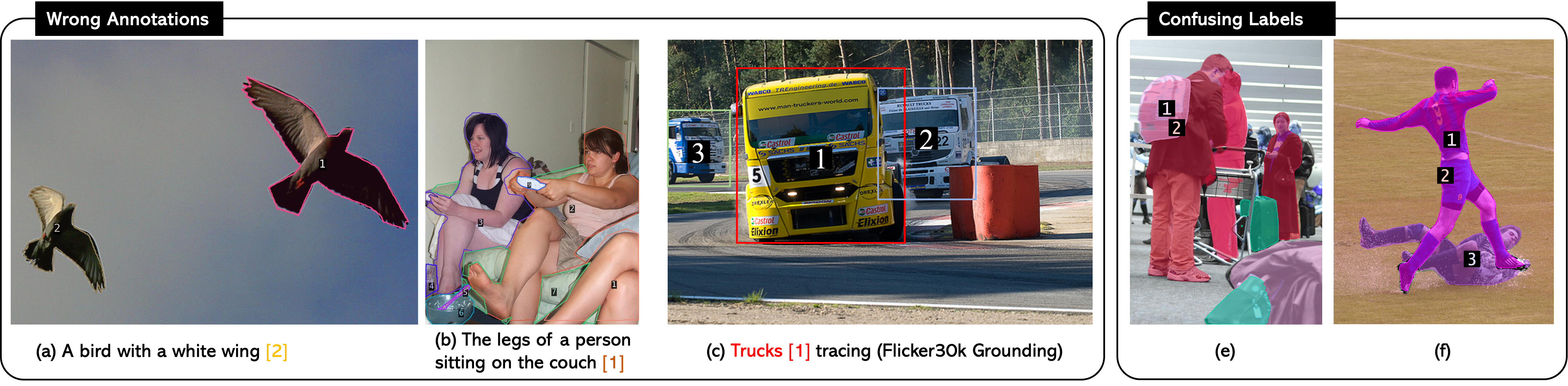}
    \caption{(a, b, c) Examples of some wrong annotations for referring segmentation (RefCOCOg) and grounding (Flikr30k). (e, f) Examples of confusing annotated labels.}
    \label{fig:error-golden-anns}       
\end{figure}

\paragraph{The ``golden'' annotations are not always golden.} When evaluating SoM, we find that a few human annotations in datasets are incorrect. Some examples are shown in Fig.~\ref{fig:error-golden-anns} (a,b,c). For example, human users annotate several people while the referring expression is ``the legs of a person sitting on the couch'' or only one truck is annotated for referring to ``truck race''. In contrast, GPT-4V with SoM yields correct answers. This implies that the ``golden'' labels in vision datasets may be subject to some noises while GPT-4V could be a good judge to help find the ambiguous annotations for further cleaning.

\paragraph{The centered marks are not always the best.} By default, our mark allocation algorithm finds the center for each region. However, we observe that putting marks at the central locations does not necessarily bring the best performance. For example in Fig.~\ref{fig:error-golden-anns} (e), there are two marks, one is for the person and the other one for the backpack. When being asked which one is the person, GPT-4V gives the mark for the backpack which is located at the upper part of the human body. A similar situation also happens in the second image where there is a background spanning the whole image. An example is shown in~\ref{fig:error-golden-anns} (f), where the center of the grassland (labeled 2) is on the person (labeled 1). These results indicate that the focus of GPT-4V to understand visual contents is not necessarily at the center. Developing a better mark allocation algorithm should be considered.

\paragraph{Dynamitc selection of mark types.} In realistic scenarios, we find dynamically determining which type of mark to use is important. For example, an image of arithmetic problems usually contains many numbers. In this case, overlaying numeric marks on to will confuse the GPT-4V. Likewise, for a screenshot of a document, it might be not rational to overlay more alphabets on the image. As a result, to fully leverage the visual capability of GPT-4V, users may need to carefully design the SoM prompts before sending the image.

\section{Related Work}

We discuss related work from the perspective of prompting.

\paragraph{LLMs and Prompting.} We have witnessed significant progress in large language models (LLMs)~\cite{brown2020language,bubeck2023sparks,chowdhery2022palm,touvron2023llama,zhang2022opt,openai2023gpt4}. Of particular, Generative Pre-trained Transformers, commonly known as GPTs~\cite{brown2020language,openai2023gpt4}, have led to a breakthrough in the realm of natural language processing. Despite the size of LLMs growing dramatically, eliciting reasoning capabilities still requires more sophisticatedly designed queries, \textit{i.e.}, prompting. In the past, a number of works attempted to do prompt engineering to endow more capability for LLMs. In-context learning is a main-stream way to teach LLMs to follow specific instructions as instantiated by a few examples~\cite{brown2020language,dong2022survey} As a follow-up, some other techniques such as chain-of-thought and tree-of-thought~\cite{wei2022chain,yao2023tree} are proposed to enhance the arithmetic, commonsense, and symbolic reasoning tasks. Analogous to these works, our \ourmodel~can be considered as a way of prompting the model to look at the image regions location-by-location. However, it also differs in that no in-context examples are needed in our method.

\paragraph{Visual and Multimodal Prompting.} 
In earlier works on interactive image segmentation~\cite{mcguinness2010comparative,chen2021conditional,chen2022focalclick}, the spatial prompting is used so that the model can take multiple user inputs (\textit{e.g.}, clicks) to gradually refine the mask. Recently, the pioneering work SAM~\cite{kirillov2023segment} and its variants~\cite{tang2023can,dai2023samaug} have proposed a unified model to support different types of prompting modes including points, boxes, and texts. In SEEM~\cite{zou2023segment}, the authors proposed a novel visual sampler to expand to visual prompting across images. Likewise, PerSAM~\cite{zhang2023personalize} and SAM-PT~\cite{rajivc2023segment} finetune SAM to support personalized segmentation and point tracking. Another line of work directly generates the prompts on the input images. In~\cite{bar2022visual}, image inpainting is used as the prompting to teach the model to predict dense outputs given example images and corresponding outputs, both in pixel space. Painter~\cite{wang2023images} and SegGPT~\cite{wang2023seggpt} share similar spirits while using masked image modeling for decoding pixels in continuous space. Visual prompting can be also applied to multimodal models. Colorful prompting tuning (CPT) is one of the first works that overlay image regions with different colors and uses masked language models to fill the blanks~\cite{yao2022cpt}. RedCircle~\cite{shtedritski2023does} draws a red circle on an image to focus the vision model on the enclosed region. In~\cite{yang2023finegrained}, the authors enhance the prompts by specifically segmenting and highlighting a target object in the image. Both methods then ask CLIP model to measure the similarity between the prompted image and a number of textual concepts.

\paragraph{LMMs and Prompting.} In a short time, large multimodal models (LMMs) have emerged as a new line of research in the community. The goal is to build a generalist multimodal system that behaves as LLMs that can perceive and reason with multimodal inputs ~\cite{liu2023visual,zhu2023minigpt,zhang2023llama,dai2023instructblip,zhang2023gpt4roi,wu2023visual,yang2023mm,chen2023shikra,li2023multimodal,Chen2023minigpt,you2023ferret}. Earlier works like MiniGPT-4~\cite{Chen2023minigpt} and LLaVa~\cite{liu2023visual} proposed a simple yet effective way to connect vision and LLMs followed by an instruction tuning. Later, a similar training strategy is applied to video tasks~\cite{lyu2023macawllm,chen2023videollm} and fine-grained vision tasks~\cite{chen2023shikra,zhang2023gpt4roi,li2023multimodal,Chen2023minigpt,you2023ferret}.
Please refer to~\cite{li2023multimodal} for a more comprehensive review of LMMs. Thus far, prompting LMMs is rarely explored in academia, partially because most of the recent open-sourced models are of limited capacity and thus incapable of such emerging capability~\cite{wei2022chain}. Most recently, GPT-4V~\cite{openai2023gpt4vision} was released followed by a comprehensive qualitative study on it~\cite{yang2023dawn}. The authors in~\cite{yang2023dawn} used a similar prompting strategy as in RedCircle~\cite{shtedritski2023does} to prompt GPT-4V. However, it only shows some referring capability of GPT-4V with hand-drawn circles or arrows, let alone a comprehensive quantitative study.

\section{Discussion}

\paragraph{The mystery in GPT-4V.} It is still mysterious why the proposed \ourmodel~could work so well with GPT-4V. Out of curiosity, we also ran some examples on other open-sourced LMMs such as LLaVa-1.5 and MiniGPT-v2. However, both models can hardly interpret the marks and ground themselves on those marks. We hypothesize a few reasons for the extraordinary visual grounding capability exhibited in GPT-4V. First, scale matters. We believe the scale of model and training data used in GPT-4V is several orders of magnitude than the aforementioned open-sourced LMMs. Second, the data curation strategy is probably another secret sauce for GPT-4V. GPT-4V could automatically associate the image regions and marks without any explicit prompt in texts. Such kind of data could be probably from literature figures, charts, \textit{etc}, which are usually clearly labeled or marked~\cite{bar2022visual}. We doubt that GPT-4V specifically employs fine-grained vision data as covered in this work. Note that the amount is extremely limited, and annotating more such kinds of data is costly, not to mention the quality is hard to control. In the end, we note that all the above suspicions are not grounded on the facts but on some empirical studies as we conducted above.

\paragraph{Connecting visual and LLMs prompting.} Despite the unknowns behind GPT-4V. Our work does take one of the first steps to connect visual prompting and LLMs prompting. In the past, many works have studied how to make vision models more promptable, which is isolated from the text promptings to LLMs. The barrier is mainly due to the language bottleneck in that we can hardly express the visual promptings precisely in language. For example, the shape/location/color of a randomly drawn stroke on the image can be hardly described verbally, unless we can encode the visual prompts and finetune the whole model~\cite{you2023ferret}. However, given the limited fine-grained training data and inferior open-sourced large vision and language models. We still see a clear gap. At the current stage, our proposed \ourmodel~demonstrates a simple yet only feasible way to inherit all existing capabilities of the strongest LMM while unleashing its strongly aspired grounding capability. We hope this work as the first to seamlessly connect visual and language prompting, could help to pave the road towards more capable LMMs.

\paragraph{Scaling data via \ourmodel{} with GPT-4V.} In the past, the whole community has strived to build fine-grained, open-vocabulary vision systems, spanning from detection~\cite{gu2021open,zhong2022regionclip,li2022grounded,zhang2022glipv2,yao2022detclip,liu2023grounding} to segmentation~\cite{ghiasi2022scaling,zou2022generalized,ding2022open}, and further expand to 3D~\cite{lu2022open,ding2023pla,liu20233d}. Though tremendous progress, we have been struggling with how to map the rich semantic knowledge from CLIP to the fine-grained domains. This is mainly due to the extremely limited number of fine-grained semantic annotations, which is still a considerable challenge to the current LMMs. As we already see, the feasibility of scaling data with fine-grained spatial has been demonstrated in SAM~\cite{kirillov2023segment} and Semantic-SAM~\cite{li2023semantic}, but how to further annotate these regions with semantic labels is still an open problem. In the light of the study in this work, we are envisioning the potential of using GPT-4V plus our \ourmodel~to intensively scale up the multimodal data which has both fine-grained spatial and detailed language description.

\section{Conclusion}
We have presented \ourmodel, a simple yet effective visual prompting mechanism for LMMs, particularly GPT-4V. 
We show that simply overlaying a number of symbolic marks on a set of regions of an input image can unleash the visual grounding ability of GPT-4V. 
We present a comprehensive empirical study on a wide range of fine-grained vision tasks to demonstrate that \textit{SoM}-prompted GPT-4V is superior to fully-finetuned specialist models and other open-sourced LMMs. 
Moreover, our qualitative results show that GPT-4V with \textit{SoM} possesses extraordinary fine-grained multimodal perception, cognition, and reasoning capabilities across the board. 
We hope that SoM would inspire future works on multimodal prompting for LMMs and pave the road towards multimodal AGI.

\section*{Acknowledgement}
We thank Fangrui Zhu for the thoughtful discussions. We thank Xuan Li and Biyi Fang from Microsoft Office Team for early brainstorming.
\bibliographystyle{plain}
\bibliography{neurips_2023}

\input{appendix/appendix}

\end{document}

%% file: tables/main_results.tex
\begin{table*}[t]
    \centering
\resizebox{0.98\linewidth}{!}{    
    \begin{tabular}{lcc|cc|cc|c|c}
    \toprule
         \multirow{2}{*}{Model} &  \multirow{2}{*}{LMM} & \multirow{2}{*}{Zero-shot} & \multicolumn{2}{c}{OV Segmentation} & \multicolumn{2}{c}{RefCOCOg} &  Phrase Grounding & VOS \\   
               &     &   & COCO & ADE20K & REC & RES & Flickr30K & DAVIS2017 \\
        \midrule
        MaskDINO~\cite{li2022mask} & \ding{55} &  \ding{55}&80.7 &-&\textcolor{gray!30}{\na}&\textcolor{gray!30}{\na}&\textcolor{gray!30}{\na}&\textcolor{gray!30}{\na} \\
        OpenSeeD~\cite{zhang2023simple} & \ding{55} &  \ding{51}&-&23.4&\textcolor{gray!30}{\na}&\textcolor{gray!30}{\na}&\textcolor{gray!30}{\na}&\textcolor{gray!30}{\na} \\       
        GLIPv2~\cite{zhang2022glipv2} & \ding{55} &  \ding{55} &-&-&-&-&$87.7^*$&\textcolor{gray!30}{\na}\\
        GDINO~\cite{liu2023grounding} & \ding{55} &   \ding{55}&\textcolor{gray!30}{\na}&\textcolor{gray!30}{\na}&$86.1^*$&\textcolor{gray!30}{\na}&90.5&\textcolor{gray!30}{\na} \\        
        X-Decoder~\cite{zou2022generalized} & \ding{55} & \ding{55}&-&-&-&$64.6^*$&\textcolor{gray!30}{\na}&62.8 \\
        PolyFormer~\cite{liu2023polyformer} & \ding{55} & \ding{55}&\textcolor{gray!30}{\na}&\textcolor{gray!30}{\na}&$85.8^*$&67.2&\textcolor{gray!30}{\na}&\textcolor{gray!30}{\na}  \\        
        SegGPT~\cite{wang2023seggpt} & \ding{55} & \ding{51}&\textcolor{gray!30}{\na}&\textcolor{gray!30}{\na}&\textcolor{gray!30}{\na}&\textcolor{gray!30}{\na}&\textcolor{gray!30}{\na}&75.6 \\
        SEEM~\cite{zou2023segment} & \ding{55} & \ding{55}&-&-&-&$65.7^*$&-&62.8 \\
        \midrule
        RedCircle~\cite{shtedritski2023does} &\ding{51} & \ding{51}&\textcolor{gray!30}{\na}&\textcolor{gray!30}{\na}&$59.4^*$&\textcolor{gray!30}{\na}&\textcolor{gray!30}{\na}&\textcolor{gray!30}{\na}\\
        FGVP~\cite{yang2023finegrained}&\ding{51} & \ding{51}&\textcolor{gray!30}{\na}&\textcolor{gray!30}{\na}& $63.3^*$ &\textcolor{gray!30}{\na}&\textcolor{gray!30}{\na}&\textcolor{gray!30}{\na}\\
        \midrule
        Shikra~\cite{chen2023shikra} &\ding{51} & \ding{55}&\textcolor{gray!30}{\na}&\textcolor{gray!30}{\na}&$82.6^*$&\textcolor{gray!30}{\na}&77.4&\textcolor{gray!30}{\na}  \\
        LLaVA-1.5~\cite{liu2023improved}  & \ding{51} & \ding{55} & \textcolor{gray!30}{\na} & \textcolor{gray!30}{\na} &63.3&\textcolor{gray!30}{\na}&\textcolor{gray!30}{\na}&\textcolor{gray!30}{\na} \\
        MiniGPT-v2~\cite{Chen2023minigpt} & \ding{51} &  \ding{55} & \textcolor{gray!30}{\na} & \textcolor{gray!30}{\na}&$84.4^*$&\textcolor{gray!30}{\na}&\textcolor{gray!30}{\na}&\textcolor{gray!30}{\na} \\
        Ferret~\cite{you2023ferret} & \ding{51} & \ding{55} & \textcolor{gray!30}{\na} & \textcolor{gray!30}{\na}& $85.8$*&\textcolor{gray!30}{\na}&81.1&\textcolor{gray!30}{\na} \\
        \midrule
        GPT-4V~\cite{openai2023gpt4vision} &  \ding{51} &  \ding{51}&\textcolor{gray!30}{\na}&\textcolor{gray!30}{\na}&25.7&\textcolor{gray!30}{\na}&\textcolor{gray!30}{\na}&\textcolor{gray!30}{\na} \\
        GPT-4V~\cite{openai2023gpt4vision} + SoM (Ours) & \ding{51} &  \ding{51}&75.7&\textbf{63.4}& \textbf{86.4}&\textbf{75.6}&89.2& \textbf{78.8} \\
    \bottomrule
    \end{tabular}    
    }
    \caption{Main quantitative results. $^*$ denotes that the number is reported in the original papers which are evaluated in the full validation datasets. Other numbers are evaluated in our sub-sampled validation datasets constraint by GPT-4V interface.}
    \label{tab:main_results}
\end{table*}

%% file: appendix/appendix.tex
\clearpage
\appendix

\section{Use Cases}

Combining GPT-4V with the proposed SoM prompting, we run many realistic use cases to check its generality and merit in comparison with the original GPT-4V.

\input{appendix/case_1}
\input{appendix/case_7}
\input{appendix/case_3}
\input{appendix/case_4}
\input{appendix/case_5}
\input{appendix/case_6}
\input{appendix/case_8}

%% file: appendix/case_1.tex
\begin{figure}[!ht]
\centering
    \begin{AIbox}{Grounded Reasoning and  Cross-Image Reference}
\begin{minipage}{0.48\textwidth}
    \includegraphics[width=1.0\textwidth]{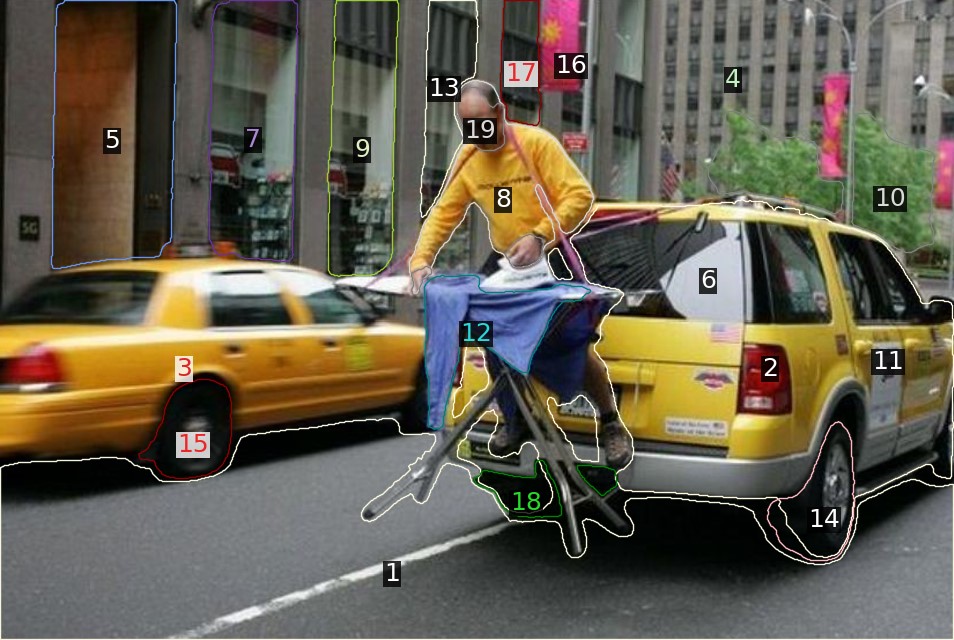}
    \\
    \\
    \footnotesize
    \textbf{\textcolor{Orange}{User}}:  Can you point me what is the most unusual thing in this image?
    \\
    \\
    \textbf{\textcolor{Green}{GPT-4V}}: 
    \\
    \\
    Certainly! The most unusual thing in the image is a man ironing clothes on an ironing board that is placed atop a moving yellow taxi cab in a city street. It's not a common sight to see someone ironing clothes on top of a taxi, especially while it's on the road!
    \\
    \\
    \textbf{\textcolor{Orange}{User}}: Can you point me what is the most unusual thing in this image?
    \\
    \\
    \textbf{\textcolor{Purple}{GPT-4V (SoM)}}: 
    \\
    \\
    The most unusual thing in this image is \textbf{the man (elements 8, 17, 19)} standing on \textbf{an ironing board (element 12)} that's placed atop a moving \textbf{yellow taxi (elements 2, 6, 11)}. He's ironing clothes in the middle of a city street, which is highly atypical and not a safe behavior. This situation stands out as the most bizarre aspect of the scene.
\end{minipage}    
\hfill
\begin{minipage}{0.48\textwidth}
    \includegraphics[width=1.0\textwidth]{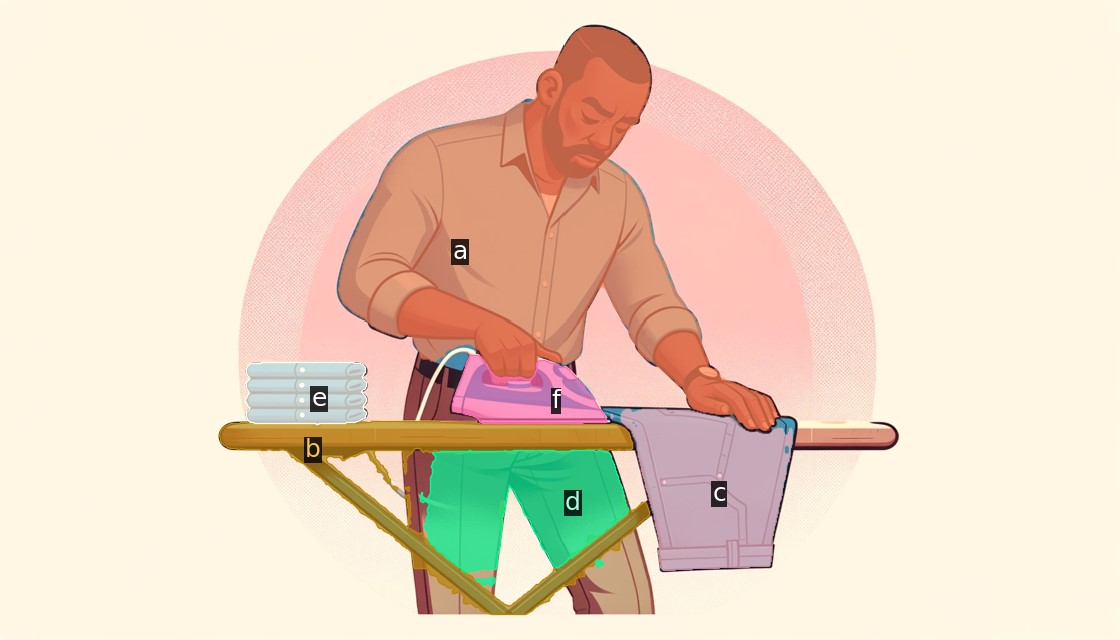}
    \\
    \\
    \footnotesize
    \textbf{\textcolor{Orange}{User}}:  What are the common objects in the two images?
    \\
    \\
    \textbf{\textcolor{Purple}{GPT-4V~(SoM)}}: The common objects in the two images are:
    \\
    \begin{itemize}[leftmargin=*]
        \item A man (In the first image, elements 8, 17, 19; In the second image, \textbf{element a})
        \item An ironing board (In the first image, element 12; In the second image, \textbf{element b})
        \item An iron (Not distinctly labeled in the first image but the man is using it; In the second image, \textbf{element f)}
        \item Clothes/shirt being ironed (Not distinctly labeled in the first image but the man is ironing it; In the second image, \textbf{element c})
    \end{itemize}
    \end{minipage}
    \end{AIbox}   
\label{fig:cross_prompt}
\caption{In comparison to GPT-4V without SoM, adding marks enables GPT-4V to ground the reasoning on detailed contents of the image (Left). Clear object cross-image references are observed on the right.}
\end{figure}

%% file: appendix/case_7.tex
\begin{figure}[t]

\centering
    \begin{AIbox}{Problem Solving}
\begin{minipage}{\textwidth}
\begin{minipage}{0.48\linewidth}
    \footnotesize
    \textbf{\textcolor{Orange}{User}}: Please solve this
    \\
    \\
    \\
    \\
    \\
\end{minipage}
\begin{minipage}{0.48\linewidth}
    \includegraphics[width=\linewidth]{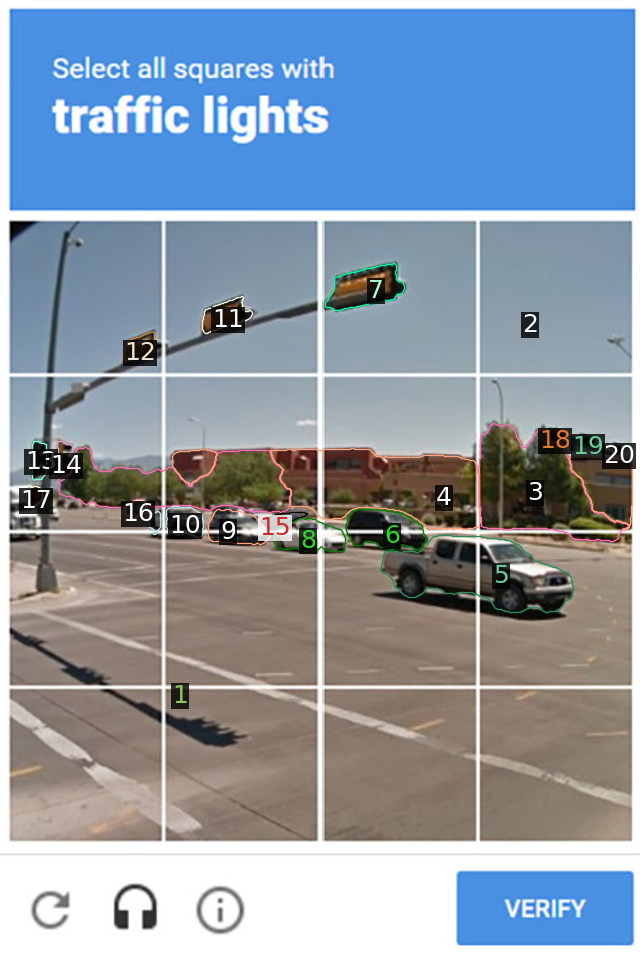}        
\end{minipage}
\end{minipage}
\begin{minipage}{\textwidth}
\textbf{\textcolor{Green}{GPT-4V}}: 
\\
\\
The squares with traffic lights in the provided CAPTCHA image are:
\begin{itemize}[leftmargin=*]
\item The first square in the first row.
\item The second square in the first row.
\item The first square in the second row.
\item The second square in the second row.
\end{itemize}

Please select these squares to solve the CAPTCHA. Remember, when you encounter CAPTCHAs in real-world scenarios, it's important to solve them yourself to ensure the security and authenticity of your online actions.
\\
\\
\textbf{\textcolor{Purple}{GPT-4V~(SoM)}}: 
\\
\\
In this image, it appears that \textbf{squares numbered 7, 11, 12} definitely contain parts of the traffic light..
\end{minipage}
\end{AIbox}   
\label{fig:case_7}
\caption{Case study on solving CAPTCHA. GPT-4V gives the wrong answer with a wrong number of squares while finding the correct squares with corresponding marks after SoM prompting.}
\end{figure}

%% file: appendix/case_3.tex
\begin{figure}

\centering
    \begin{AIbox}{Knowledge Sharing}
\begin{minipage}{\textwidth}
\begin{minipage}{0.38\linewidth}
    \footnotesize
    \textbf{\textcolor{Orange}{User}}: What is in this dish?
    \\
    \\
    \\
    \\
    \\
\end{minipage}
\begin{minipage}{0.58\linewidth}
    \includegraphics[width=\linewidth]{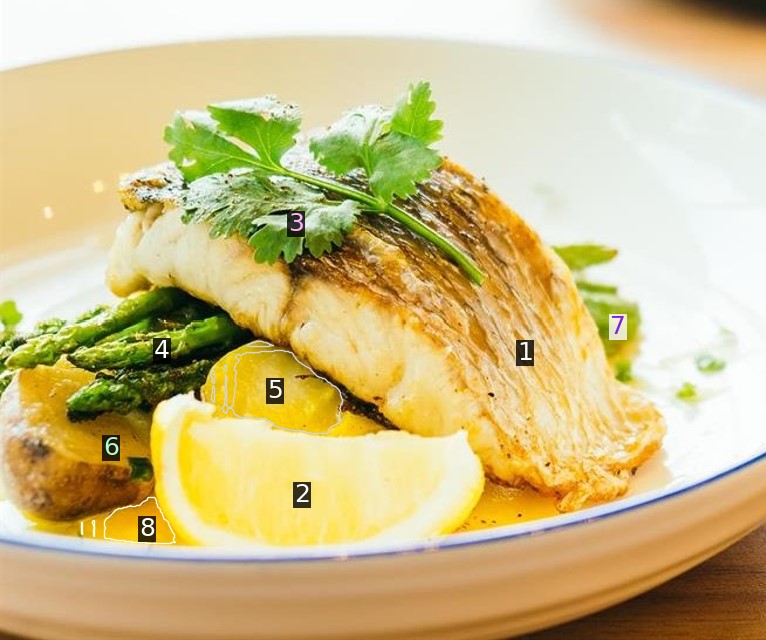}        
\end{minipage}
\end{minipage}
\begin{minipage}{\textwidth}
\textbf{\textcolor{Green}{GPT-4V}}: 
\\
\\
The dish appears to feature:
\begin{itemize}[leftmargin=*]
\item A grilled or pan-seared fish fillet, which looks like it could be a type of white fish, possibly sea bass, cod, or halibut.
\item Asparagus spears.
\item Potatoes, possibly roasted or boiled.
\item A wedge of lemon for squeezing over the dish.
\item Garnished with fresh parsley or cilantro leaves.
\item There might be a light sauce or seasoning drizzled over the fish and vegetables, but it's hard to specify its ingredients from the image alone.
\end{itemize}

This looks like a delicious and healthy meal!
\\
\\
\textbf{\textcolor{Purple}{GPT-4V~(SoM)}}: 
\\
\\
Here's a description of the items in the dish:
\begin{enumerate}[leftmargin=*]
    \item \textbf{Fish Fillet}: The main component of the dish appears to be a fillet of fish, likely grilled or pan-seared.
    \item \textbf{Lemon Wedge}: A common accompaniment with fish dishes to enhance flavor when squeezed over the top.
    \item \textbf{Parsley or Cilantro}: This green herb is used as a garnish and can add a fresh flavor to dishes.
    \item \textbf{Asparagus}: These green stalks are often steamed, sautéed, or grilled and served as a side.
    \item \textbf{Sauce or Broth}: There is a yellowish sauce or broth on the plate.
    \item \textbf{Potato}: It looks like a roasted or boiled potato.
    \item \textbf{Garnish}: These could be green onions or chives scattered on the plate to add color and flavor.
    \item \textbf{Sauce or Oil Drizzle}: There's a darker drizzle around the potato, which could be a type of sauce or perhaps a balsamic reduction.
\end{enumerate}
The dish seems to be a well-balanced meal with a protein (fish), vegetables (asparagus and potato), and various flavor enhancers (sauces, lemon, and herbs).
\\
\end{minipage}
\end{AIbox}   
\label{fig:case_3}
\caption{Case study on an image of dish for GPT-4V. GPT-4V does not produce a grounded answer with the original image. Based on SoM prompting, GPT-4V not only speaks out the ingredients but also corresponds them to the regions.}
\end{figure}

%% file: appendix/case_4.tex
\begin{figure}

\centering
    \begin{AIbox}{Personalized Suggestion}
\begin{minipage}{\textwidth}
\begin{minipage}{0.38\linewidth}
    \footnotesize
    \textbf{\textcolor{Orange}{User}}: I am on a diet, what are the top things I should avoid eating?
    \\
    \\
    \\
    \\
    \\
\end{minipage}
\hfill
\begin{minipage}{0.58\linewidth}
    \includegraphics[width=\linewidth]{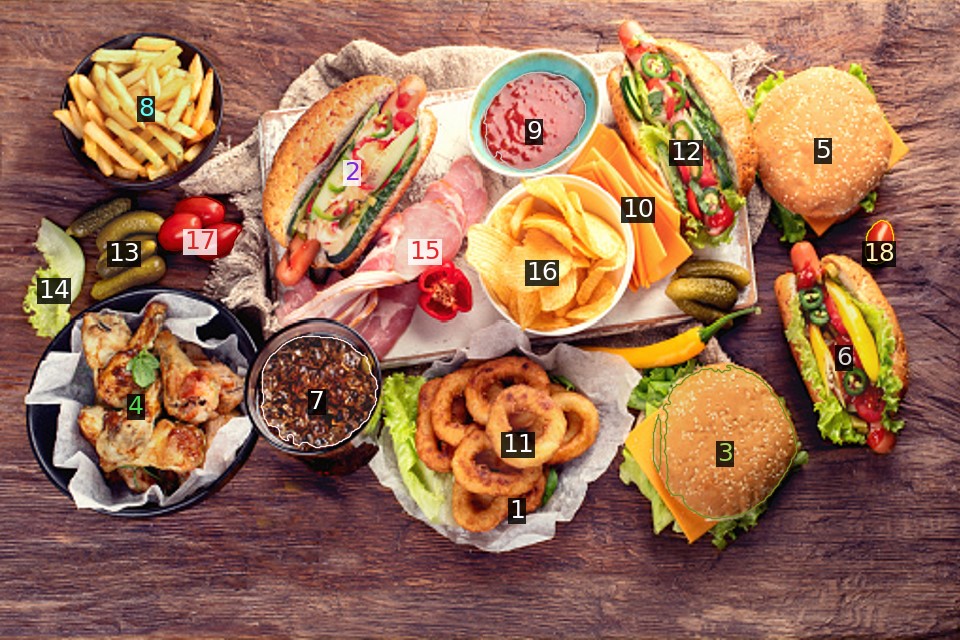}        
\end{minipage}
\end{minipage}
\begin{minipage}{\textwidth}
\textbf{\textcolor{Green}{GPT-4V}}: 
\\
\\
If you're on a diet, it would depend on the specific type of diet you're following. However, based on the image you provided, here are some common items that many diets might limit or avoid:
\begin{itemize}[leftmargin=*]
\item Burgers: Especially if made with high-fat meats or served with cheese and mayonnaise.
\item Fries: Deep-fried foods are typically high in calories and unhealthy fats.
\item Chips: Can be high in salt and fats, especially if they're commercially produced.
\item Onion rings: Another deep-fried item.
\item Sandwiches with white bread: Many diets recommend whole grains over refined grains.
\item Processed meats: Like the ham in the sandwich, as they can be high in sodium and preservatives.
\item Sauces: Some can be high in sugar, sodium, or unhealthy fats.
\item Soft drinks or sweetened beverages: High in sugars and empty calories.
\end{itemize}

Always consult with a nutritionist or dietician to get guidance tailored to your specific needs and health goals. They can provide recommendations based on your diet and personal health objectives.
\\
\\
\textbf{\textcolor{Purple}{GPT-4V~(SoM)}}: 
\\
\\
Based on the typical diet considerations, here are the top items from the provided image that you should most likely avoid or limit:
\begin{enumerate}[leftmargin=*]
\item \textbf{{Onion rings (1 and 11)}} - Fried foods like these are often high in unhealthy fats and calories.
\item \textbf{{Fries (8)}} - Another fried item that can be calorie-dense and contain unhealthy fats.
\item \textbf{{Burgers (3 and 5)}} - Depending on their composition, they can be high in calories, unhealthy fats, and may contain processed meats.
\item \textbf{{Processed meats (2 and 15)}} - These can be high in sodium, unhealthy fats, and other additives.
\item \textbf{{Chips (16)}} - Typically high in unhealthy fats and can be calorie-dense.
\end{enumerate}
While indulging occasionally is okay, it's best to consume these items in moderation if you're trying to maintain a healthy diet.
\\
\end{minipage}
\end{AIbox}   
\label{fig:case_4}
\caption{SoM-pormpted GPT-4V gives very precise suggestions while the original one fails, even with hallucinated foods, \textit{e.g.}, soft drinks.}
\end{figure}

%% file: appendix/case_5.tex
\begin{figure}

\centering
    \begin{AIbox}{Tool Usage Instruction}
\begin{minipage}{\textwidth}
\begin{minipage}{0.38\linewidth}
    \footnotesize
    \textbf{\textcolor{Orange}{User}}: How to use this controller?
    \\
    \\
    \\
    \\
    \\
\end{minipage}
\hfill
\begin{minipage}{0.58\linewidth}
    \includegraphics[width=\linewidth]{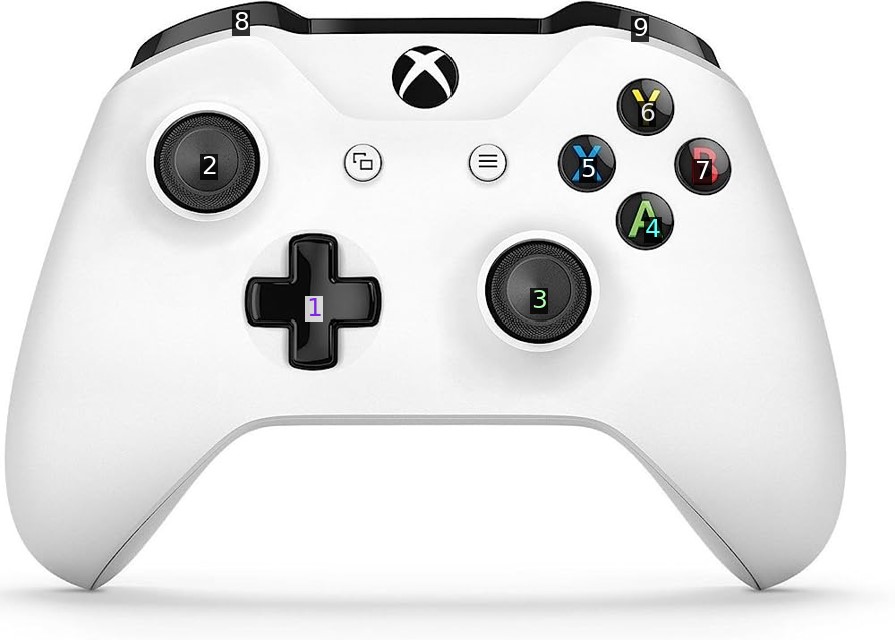}        
\end{minipage}
\end{minipage}
\begin{minipage}{\textwidth}
\textbf{\textcolor{Purple}{GPT-4V~(SoM)}}: 
\\
\\
\footnotesize
This appears to be an Xbox One or Xbox Series X|S controller. Here's a basic rundown on how to use it:

\begin{itemize}[leftmargin=*]
\item \textbf{Thumbsticks (2 \& 3)}: These are primarily used for movement and aiming. 
    \begin{itemize}
        \item \textbf{Left Thumbstick (2)}: Controls the movement of the character.
        \item \textbf{Right Thumbstick (3)}: Controls the camera or aiming reticle.
    \end{itemize}
\item \textbf{Directional Pad (D-Pad) (1)}: Used for various functions such as navigating menus, selecting weapons, or other in-game actions.
\item \textbf{Face Buttons (4, 5, 6, \& 7)}:
    \begin{itemize}
        \item \textbf{A (4)}: Typically the main action or "accept" button.
        \item \textbf{B (7)}: Often the "back" or "cancel" button.
        \item \textbf{X (5)}: Varies by game, frequently used for interactions.
        \item \textbf{Y (6)}: Like X, its function varies but is commonly used for secondary actions or interactions.
    \end{itemize}
\item \textbf{Bumpers and Triggers (8 \& 9)}:
    \begin{itemize}
        \item \textbf{Left and Right Bumpers (8)}: Used for quick actions or toggling between items.
        \item \textbf{Left and Right Triggers}: Used for primary actions like shooting or accelerating in racing games.
    \end{itemize}
\item {Menu and View Buttons}: Located in the center, used to bring up in-game menus or other overlays.
\item {Xbox Button}: Press to go to the Xbox home or turn the controller on/off. Long press brings up a menu to turn off the controller or console.
\item {Pairing Button}: Located on the top edge, used for pairing the controller.
\item {3.5mm Audio Jack}: On the bottom, plug in headphones or a headset.
\item {USB Port}: On the top edge, used for charging or wired play.
\end{itemize}

Refer to the game's manual or in-game control settings for exact functions for each game.
\\
\end{minipage}
\end{AIbox}   
\label{fig:case_5}
\caption{Likewise, GPT4-V with SoM can help to provide thorough tool usage instruction, teaching users the function of each button on a controller. Note that this image is not fully labeled, while GPT-4V can also provide information about the non-labeled buttons.}
\end{figure}

%% file: appendix/case_6.tex
\begin{figure}[t]

\centering
    \begin{AIbox}{2D Game Planning}
\begin{minipage}{\textwidth}
\begin{minipage}{0.38\linewidth}
    \footnotesize
    \textbf{\textcolor{Orange}{User}}: How can I get the coins on top?
    \\
    \\
    \\
    \\
    \\
\end{minipage}
\hfill
\begin{minipage}{0.58\linewidth}
    \includegraphics[width=\linewidth]{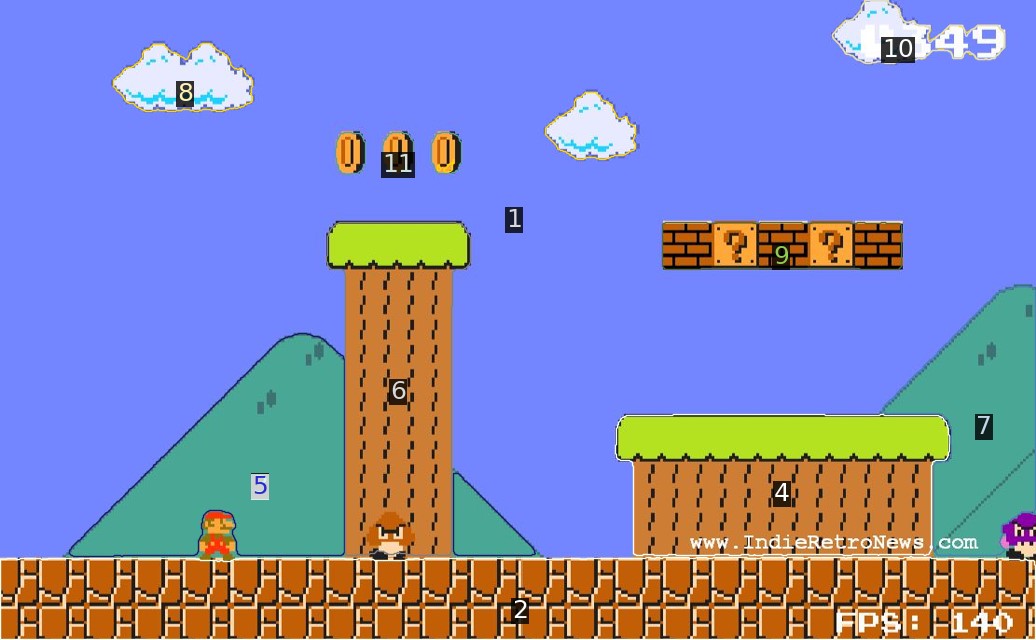}        
\end{minipage}
\end{minipage}
\begin{minipage}{\textwidth}
\textbf{\textcolor{Purple}{GPT-4V~(SoM)}}: 
\\
\\
\footnotesize
It looks like you're playing a platformer game reminiscent of Super Mario Bros. Here's a general strategy based on the layout:

\begin{itemize}[leftmargin=*]
\item {Elevated Platforms}: \textbf{The platforms labeled as '4' and '7'} are the key. You need to first get onto these platforms to reach greater heights. Depending on the game mechanics, you may need to find a run-up or utilize a jump-boosting power-up if available.

\item {Bricks and Blocks}: \textbf{The bricks labeled as '9'} might contain some power-ups or they might be breakable. In classic platformers like Super Mario, hitting such blocks from below can reveal power-ups or even create platforms. \textbf{The same goes for the block labeled as '1'}.

\item {Vertical Jump}: \textbf{Once on the elevated platform '4'}, try to perform a vertical jump to reach \textbf{the coins labeled '11'}. Ensure that you time your jump correctly and use any speed or jump boosts, if available.

\item {Momentum}: Some games also allow for gaining extra height in jumps by building momentum. If this game works similarly, try running from a distance and jumping at the last moment to grab the coins.

\item {Look for Hidden Power-ups}: Some games hide power-ups in the level that can assist players in reaching difficult places. These might be hidden in unmarked bricks or in other places. If you have abilities like a double jump, use them.

\item {Game Mechanics}: Every game can have unique mechanics. If there are any other controls or mechanics you haven't tried, consider experimenting with them. Maybe there's a way to bounce, double jump, or use an enemy to boost your jump height.    
\end{itemize}
If you've tried all the general strategies and still can't reach the coins, it might be beneficial to look up specific strategies or walkthroughs for the particular game you're playing. Good luck!
\\
\end{minipage}
\end{AIbox}   
\label{fig:case_6}
\caption{GPT-4V with SoM gives a reasonable suggestion on how to achieve a goal in a gaming scenario.}
\end{figure}

%% file: appendix/case_8.tex
\begin{figure}[t]
\centering
\begin{AIbox}{Simulated Navigation}
\begin{minipage}{0.48\textwidth}
    \includegraphics[width=1.0\textwidth]{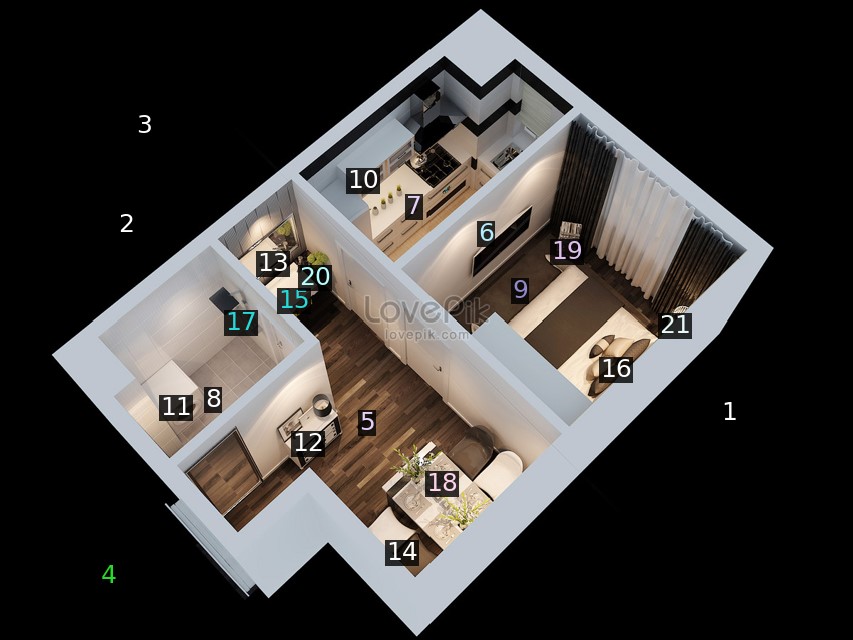}
    \\
    \\
    \footnotesize
    \textbf{\textcolor{Orange}{User}}: I am a bit hungry, where can I find some food?
    \\
    \\
    \textbf{\textcolor{Purple}{GPT-4V (SoM)}}: 
    \\
    Based on the image provided, it looks like you're viewing a floor plan of an apartment. For food, \textbf{you should head to the kitchen area which is labeled with the number "10"}. That's where you're most likely to find food or kitchen appliances to prepare a meal. If you're in this apartment, head to that section! If you're not, consider checking your own kitchen or a nearby restaurant. Enjoy your meal!
    \\
    \\
    \textbf{\textcolor{Orange}{User}}: cool, where should I sit to eat if I find some food from 10?
    \\
    \\
    \textbf{\textcolor{Purple}{GPT-4V (SoM)}}: 
    \\
    Based on the floor plan, the most suitable place to sit and eat would be at \textbf{the dining area labeled with the number "14". It appears to have a dining table and chairs}, making it a comfortable spot for enjoying a meal.
    \\
    \\
    \textbf{\textcolor{Orange}{User}}: ok, i got a dirty hand, what should I do?
    \\
    \\
    \textbf{\textcolor{Purple}{GPT-4V (SoM)}}: 
    \\
    Based on the floor plan, you should head to the bathroom areas to wash your hands. \textbf{You can go to the space labeled with the number "13" which appears to be a bathroom}. It's likely to have a sink and soap where you can clean your hands. Always remember to wash your hands thoroughly, especially before eating!  
\end{minipage}
\hfill
\begin{minipage}{0.48\linewidth}
    \textbf{\textcolor{Orange}{User}}: now I need to get some sleep.
    \\
    \\
    \textbf{\textcolor{Purple}{GPT-4V (SoM)}}: 
    \\
    For sleep, you should \textbf{head to the area labeled with the number "19"}. It appears to be a bedroom with a bed, which would be the most suitable place for rest. Have a good sleep!
    \\
    \\
    \textbf{\textcolor{Orange}{User}}: where is the TV, I want to turn it on.
    \\
    \\
    \textbf{\textcolor{Purple}{GPT-4V (SoM)}}: 
    \\
    \textbf{The TV is located in the area labeled with the number "6"}. You can head over there to turn it on and enjoy some entertainment.
    \\
    \\
    \textbf{\textcolor{Orange}{User}}: Enough to watch a TV, I want to take a shower.
    \\
    \\
    \textbf{\textcolor{Purple}{GPT-4V (SoM)}}: 
    \\
    To take a shower, you should \textbf{head to the space labeled with the number "17"}. It appears to be a shower area in the bathroom. Make sure you have your toiletries and towel ready, and enjoy your shower!
    \\
    \\
    \textbf{\textcolor{Orange}{User}}: okay, need to wash some clothes, put them where?
    \\
    \\
    \textbf{\textcolor{Purple}{GPT-4V (SoM)}}: 
    \\
    For washing clothes, you should \textbf{head to the area labeled with the number "8"}. It appears to be a location designated for a washing machine or laundry area. You can place your clothes there and start the washing process.
\end{minipage}
\end{AIbox}   
\label{fig:case_8}
\caption{GPT-4V with SoM has the potential to help embodied navigation in that it can quickly help to find the regions/locations of particular functionality.}
\end{figure}